\documentclass[11pt]{article}

\usepackage[final]{acl}

\usepackage{times}
\usepackage{latexsym}
\usepackage{textcomp}
\usepackage{upquote}
\usepackage[T1]{fontenc}
\usepackage[utf8]{inputenc}
\usepackage{inconsolata}
\usepackage{microtype}
\usepackage{graphicx}
\graphicspath{{figures/}}
\usepackage{amsmath,amssymb,amsfonts}
\usepackage{subcaption}
\usepackage{multirow}
\usepackage{pifont}
\usepackage{listings}
\usepackage{xcolor}
\usepackage{colortbl}
\usepackage{booktabs}
\usepackage{algorithm}
\usepackage{algpseudocode}
\usepackage{float}
\usepackage{amsthm}
\newtheorem{theorem}{Theorem}
\newcommand{\yes}{\ding{51}}
\newcommand{\no}{\ding{55}}
\newcommand{\eg}{{\it e.g.}}

\title{Linguistics-Aware Non-Distortionary LLM Watermarking}

\author{
  Shinwoo Park$^1$\thanks{This work was carried out while the author was at Yonsei University.},
  Hyejin Park$^2$,
  Hyeseon An$^3$,
  \and
  Yo-Sub Han$^3$\thanks{Corresponding Author.} \\
  $^1$University of Luxembourg \\
  $^2$Rensselaer Polytechnic Institute \\
  $^3$Yonsei University \\
  \href{mailto:shinwoo.park@uni.lu}{\texttt{shinwoo.park@uni.lu}}, \\
  \href{mailto:parkh12@rpi.edu}{\texttt{parkh12@rpi.edu}}, \\
  \texttt{\{\href{mailto:hsan@yonsei.ac.kr}{hsan},
  \href{mailto:emmous@yonsei.ac.kr}{emmous}\}@yonsei.ac.kr}
}

\begin{document}

\maketitle

\begin{abstract}
Watermarking should identify language-model output without degrading 
quality or limiting verification to the model provider.
Multilingual deployment makes this harder because morphology, segmentation, 
and script change where watermark evidence can be naturally embedded.
We introduce LUNA, a linguistically adaptive watermark that combines 
model-free detection with single-token non-distortion under the standard 
random-key model.
LUNA estimates normalized next-tag entropy from part-of-speech contexts in 
an external corpus and uses it to set the depth of a non-distortionary 
binary tournament sampler; the detector reconstructs the same schedule from 
text, a tokenizer, a tagger, and a secret key.
We evaluate LUNA on six typologically diverse languages and two domains against 
eight primary baselines.
LUNA attains an AUROC of $0.9959$ and the lowest mean absolute median perplexity 
shift, $0.045$, across the $12$ settings; its $95\%$ bootstrap interval 
$[0.022, 0.073]$ lies below all baseline intervals.
LUNA also records the lowest mean on Self-BLEU, Distinct-1, surprisal, and 
entropy shifts; it is the only method that simultaneously achieves AUROC 
$>0.99$ and $|\Delta\mathrm{PPL}_{\mathrm{med}}|<0.1$ in a majority of 
settings, reaching this regime in $9$ of the $12$ settings while no baseline 
reaches it in more than $2$.
Our code is available at \url{https://github.com/Shinwoo-Park/luna_watermark}.
\end{abstract}

\section{Introduction}
\label{sec:intro}

Large language models now generate fluent text at scale, creating practical 
needs for provenance, attribution, and disinformation 
control~\citep{liu2024survey,lalai-etal-2025-intentions,euaiact2024}.
Decoding-time watermarking addresses these needs by embedding a statistical 
signal during generation and testing for it after 
deployment~\citep{kirchenbauer2023watermark,dathathri2024scalable}.
A deployment-ready watermark should satisfy three properties together: 
\textbf{single-token non-distortion}, where the next-token distribution 
equals the base distribution after marginalizing over watermark 
randomness~\citep{aaronson2022watermark,kuditipudi2024robust,dathathri2024scalable}; 
\textbf{model-free detection}, so platforms and third-party auditors can 
verify provenance without querying the original model or a 
surrogate~\citep{kirchenbauer2023watermark,park2026linguistics}; 
and \textbf{adaptivity}, since different contexts provide different amounts 
of reliable 
capacity~\citep{lu-etal-2024-entropy,wang-etal-2025-morphmark,park2026linguistics}.
Prior work has not, to our knowledge, combined all three; 
recent adaptive non-distortionary 
designs
draw adaptivity from model-side 
uncertainty, which ties detection to logits or 
surrogate forward passes and 
weakens public verifiability.

The central observation behind LUNA is linguistic.
Languages differ systematically in how much grammatical choice a position 
permits.
For example, after the part-of-speech context 
\texttt{DET ADJ} in English~(\eg\ ``a quiet ...''), 
the next tag is almost always \texttt{NOUN}, 
carrying little grammatical choice; 
after the Korean morpheme sequence 
\texttt{NNG JKO}~(object marker), the 
next slot can be a verb, adverbial, 
or adnominal modifier, spreading probability 
over several tags.
The first context yields a low normalized 
next-tag entropy, the second a 
high one.
Such variation reflects the language and 
its analysis pipeline rather 
than any particular language model, 
so a part-of-speech tagged corpus 
can estimate a reusable signal of 
local 
syntactic uncertainty~\citep{comrie1989language,greenberg1963some,haspelmath2005world}.
Paired with a prefix-measurable 
non-distortionary sampler, this signal 
guides watermark capacity toward positions 
with greater grammatical choice 
while preserving the one-step marginal distribution, 
and it enables 
detection from the tokenizer, a tagger, 
and the secret key without model 
logits.

We introduce 
LUNA~(\textbf{L}ing\textbf{u}istics-Aware \textbf{N}on-Distortionary LLM W\textbf{a}termarking).
LUNA estimates normalized next-tag entropy 
for part-of-speech contexts from 
an external corpus, 
reconstructs the current context $c_t$ from the prefix, 
retrieves $\lambda(c_t)\in[0,1]$, 
and maps it to a depth $m_t$ for a binary 
tournament sampler~\citep{dathathri2024scalable}.
The schedule is prefix-measurable because $m_t$ 
is fixed before sampling 
$x_t$, which preserves single-token marginals 
under the random-key model 
and allows the detector to reconstruct
the same depth sequence from text alone.
We evaluate LUNA on a compact, 
typology-aware six-language grid 
spanning 
analytic English~\citep{quirk1985comprehensive,marcus1993building}, 
isolating Chinese~\citep{li-thompson-1981,xue2005penn}, 
agglutinative Korean~\citep{sohn1999korean,kim2024does} 
and Japanese~\citep{tsujimura1996introduction,kuno1973structure}, 
fusional German~\citep{haider2010syntax,vikner1995verb}, 
and 
templatic Semitic 
Arabic~\citep{mccarthy1981prosodic,watson2002phonology,ryding2005reference}.
Across the multilingual grid,
LUNA retains near-saturated detection
while occupying the low-shift end of the
measured detection--quality trade-off.

\section{Related Work}
\label{sec:related-work}

\begin{table*}[ht!]
\centering
\small
\begin{tabular}{@{}l c c c c c@{}}
\toprule
\textbf{Method} 
& \textbf{Single-token} 
& \textbf{Adaptive} 
& \textbf{Adaptive} 
& \textbf{Model-free} 
& \textbf{Linguistic} \\
& \textbf{Non-distortion} 
& \textbf{Insertion} 
& \textbf{Detection} 
& \textbf{Detection} 
& \textbf{Signal} \\
\midrule
KGW~\citep{kirchenbauer2023watermark}        & \no  & \no  & \no  & \yes & \no \\
EWD~\citep{lu-etal-2024-entropy}             & \no  & \no  & \yes & \no  & \no \\
SWEET~\citep{lee-etal-2024-wrote}            & \no  & \yes & \yes & \no  & \no \\
MorphMark~\citep{wang-etal-2025-morphmark}   & \no  & \yes & \no  & \yes & \no \\
STELA~\citep{park2026linguistics}            & \no  & \yes & \yes & \yes & \yes \\
GumbelSoft (diversified)\textsuperscript{$\dagger$}~\citep{fu-etal-2024-gumbelsoft}   & \no$^{\dagger}$ & \no & \no & \yes & \no \\
EXP~\citep{aaronson2022watermark}            & \yes & \no  & \no  & \yes & \no \\
SynthID-Text~\citep{dathathri2024scalable}   & \yes & \no  & \no  & \yes & \no \\
\midrule
\textbf{LUNA}                                 & \textbf{\yes} & \textbf{\yes} & \textbf{\yes} & \textbf{\yes} & \textbf{\yes} \\
\bottomrule
\end{tabular}
\caption{Operational taxonomy of the primary baselines 
and LUNA. 
Column definitions appear in Section~\ref{sec:rw-taxonomy}. 
The dagger~($\dagger$) marks the diversified GumbelSoft variant, 
which softens the deterministic Gumbel-max decoding and therefore does not 
inherit the exact single-token distribution-preservation guarantee of EXP or 
SynthID-Text.}
\label{tab:taxonomy}
\end{table*}

\subsection{Distribution-Shifting and Adaptive Watermarks}
\label{sec:rw-distortionary}

The first family of language-model watermarks 
embeds detectable evidence by modifying 
the next-token distribution during decoding.
KGW~\citep{kirchenbauer2023watermark} partitions 
the vocabulary into keyed green and red lists, 
biases green-list logits before sampling, 
and detects the watermark through a 
one-proportion test on the observed green-token count.
This design enables efficient 
model-free detection because the detector 
needs the text, key, and tokenizer 
rather than target-model logits.
The same mechanism makes KGW 
single-token distortionary, 
since the sampler explicitly changes 
the probability mass assigned to green-list tokens.

Adaptive variants change insertion or 
detection across positions.
SWEET~\citep{lee-etal-2024-wrote} 
targets code generation and applies KGW-style bias only 
at positions whose model entropy exceeds a threshold; 
its detector reuses the same threshold.
EWD~\citep{lu-etal-2024-entropy} leaves KGW-style 
generation unchanged and instead weights detected 
tokens by model-side entropy.
MorphMark~\citep{wang-etal-2025-morphmark} 
adapts insertion strength according to the 
cumulative probability mass of green-list tokens and 
keeps KGW-style detection.
STELA~\citep{park2026linguistics} estimates 
part-of-speech context indeterminacy from a corpus 
and uses that signal to modulate both green-list 
bias and detection weighting.
These methods show that context-dependent allocation 
can improve watermarking, while their operational 
requirements differ: SWEET and EWD require 
model-side entropy at detection time, 
MorphMark preserves KGW-style model-free detection, 
and STELA obtains model-free linguistic adaptivity through 
a tagger rather than logits.

\subsection{Distribution-Preserving and Gumbel-Based Watermarks}
\label{sec:rw-preserving} 

The second family seeks watermark evidence 
while preserving the base decoding 
distribution under explicit randomness assumptions.
Aaronson-style 
exponential-minimum sampling~\citep{aaronson2022watermark} and 
the framework of 
\citet{kuditipudi2024robust} instantiate 
this principle through 
keyed sampling schemes 
such as inverse-transform and 
exponential-minimum sampling.
SynthID-Text~\citep{dathathri2024scalable} 
introduces tournament sampling 
and supports 
a single-token non-distortionary configuration with 
binary tournaments; 
its detector computes keyed scores without 
using the language model at detection time.
Although DAWA~\citep{he2025theoretically} 
jointly optimizes generation and detection under 
explicit distortion constraints, 
its adaptive mechanism 
is derived from the model distribution and a 
surrogate model rather than from 
external linguistic signals.
GumbelSoft~\citep{fu-etal-2024-gumbelsoft} 
addresses generation diversity in Gumbel-keyed watermarking.
It replaces deterministic decoding with a softmax variant of 
Logits-Addition, sampling from 
\(\mathrm{softmax}((\ell_t+\xi_t)/\tau)\), and detects by 
aggregating keyed scores \(\xi_t[x_t]\) for observed tokens.
This makes GumbelSoft a strong model-free baseline, 
although the paper does not establish the exact one-step 
distribution-preservation guarantee that we assign to 
EXP~\citep{aaronson2022watermark} and 
the non-distortionary SynthID-Text configuration in 
Table~\ref{tab:taxonomy}.

\subsection{Multilingual and Cross-Lingual Watermarks}
\label{sec:rw-multilingual}

Multilingual and cross-lingual settings 
expose difficulties that English-only 
evaluations can hide: 
translation, segmentation, morphology, and script 
can alter the evidence available to a detector.
Prior work examines watermark survival under 
translation, cross-lingual manipulation, and 
back-translation robustness~\citep{he-etal-2024-watermarks,al-ghanim-etal-2025-evaluating,mohamed2025multilingual}, 
robustness studies show that paraphrasing, editing, and other
transformations can substantially change watermark
evidence~\citep{rastogi-pruthi-2024-revisiting,liang-etal-2025-watermark},
and benchmark suites evaluate watermarks jointly on detection and
generation quality~\citep{tu-etal-2024-waterbench}.

This line of work primarily asks whether watermark evidence remains detectable 
after text has been transformed across languages, domains, or surface forms.
LUNA addresses a complementary question at generation time: 
where should watermark capacity enter the text when languages differ in 
morphology, segmentation, word order, and script?
Its schedule conditions tournament depth on language-specific 
part-of-speech context entropy, making the source of watermark evidence 
measurable before any downstream transformation occurs.

\subsection{Operational Taxonomy}
\label{sec:rw-taxonomy}

Table~\ref{tab:taxonomy} summarizes 
the primary baselines and LUNA.
\texttt{Single-token Non-distortion} 
denotes one-step marginal preservation under the
stated sampling assumptions; 
\texttt{Adaptive Insertion} and 
\texttt{Adaptive Detection} 
denote context-dependent signal allocation during generation and detection; 
\texttt{Model-free Detection} 
denotes detection without target or surrogate 
language-model forward passes; and 
\texttt{Linguistic Signal} 
denotes whether the 
adaptive signal is derived from corpus-estimated linguistic structure rather 
than model logits.

Green-list methods obtain evidence through logit bias and sacrifice 
single-token non-distortion.
Distribution-preserving methods preserve 
one-step marginals under their sampling assumptions, yet they do not use an 
interpretable linguistic signal.
Adaptive methods split across insertion and detection, with some relying on 
model-side entropy.
LUNA occupies the missing operational point: it inherits a non-distortionary 
tournament backbone, replaces fixed schedules with 
part-of-speech context uncertainty, adapts both insertion 
and detection through the same signal, and supports 
detection without target or surrogate model access.

\section{Background}
\label{sec:background}

\subsection{Typological Stress Test}
\label{sec:bg-typology}

LUNA assumes that watermark capacity 
should track how much grammatical 
choice a position affords; 
this depends on the morphological and syntactic 
profile of the language.
The evaluation uses six languages that 
stress distinct interactions among 
morphology, word order, spacing, and script: 
analytic English and 
isolating Chinese~(low-inflection SVO with different 
writing systems), 
agglutinative Korean and 
Japanese~(particles and endings creating 
fine-grained POS sequences), 
fusional German~(verb-second syntax with case and agreement), 
and 
templatic Arabic~(Semitic root-and-pattern morphology with an abjad 
script).
Table~\ref{tab:typology-short} summarizes the stress points.

\begin{table}[t]
\centering
\small
\begin{tabular}{@{}l l l@{}}
\toprule
\textbf{Lang.} & \textbf{Profile} & \textbf{Stress point} \\
\midrule
EN & Analytic & SVO, light inflection \\
ZH & Isolating & No spacing, particles \\
KO & Agglutinative & Morphemes, particles \\
JA & Agglutinative & Segmentation, mixed script \\
DE & Fusional & Case, Verb-Second~(V2) syntax \\
AR & Templatic & Abjad, root-pattern morphology \\
\bottomrule
\end{tabular}
\caption{Typological stress test used by the evaluation.}
\label{tab:typology-short}
\end{table}

\subsection{Tournament Sampling and Detection}
\label{sec:bg-tournament}

SynthID-Text is a generative watermarking scheme 
built from three components: a random seed generator, 
a sampling algorithm, and a 
scoring function.
Let \(\mathcal{V}\) denote the vocabulary, 
\(x_{<t}\) the prefix before position \(t\), 
and
\[
p_t(v)=\Pr_{\mathrm{base}}(x_t=v\mid x_{<t})
\]
the next-token distribution passed to the sampling layer.
Given a seed \(r_t\) derived from the recent context 
and a watermarking key, 
SynthID-Text defines layer-wise keyed functions \(g_1,\ldots,g_m\).
For the binary configuration used in the 
non-distortionary setting, each \(g_\ell(v,r_t)\) 
assigns a value in \(\{0,1\}\) to candidate token \(v\).

At a fixed depth \(m\), tournament sampling 
first draws \(2^m\) candidate tokens from \(p_t\), 
with repetitions allowed.
It then runs an \(m\)-layer knockout tournament: 
layer \(\ell\) compares paired candidates with \(g_\ell(\cdot,r_t)\), breaks ties randomly, 
and passes winners to the next layer until 
one token remains.
SynthID-Text also admits 
a distortionary configuration with more than 
two competitors per match, 
which strengthens the watermark at the cost of 
token-level distortion.
This subsection uses only 
the fixed-depth binary configuration; 
Section~\ref{sec:method} introduces 
the adaptive depth schedule used by LUNA.

For detection, SynthID-Text recomputes 
the same keyed scores on an observed sequence 
and aggregates them into a text-level statistic.
For fixed depth \(m\), 
a simplified score over valid positions \(\mathcal{I}\) is
\begin{equation}
\operatorname{Score}_m(x)=
\frac{1}{m|\mathcal{I}|}
\sum_{t\in\mathcal{I}}\sum_{\ell=1}^{m}g_\ell(x_t,r_t).
\label{eq:synthid-score}
\end{equation}
Watermarked text tends to receive higher 
keyed scores because tournament sampling 
favors candidates with larger layer values.
This score depends on the observed text, 
the key, and the seed generator; 
it does not require a forward pass through 
the language model at detection time.

\begin{figure*}[ht!]
\includegraphics[width=\textwidth]{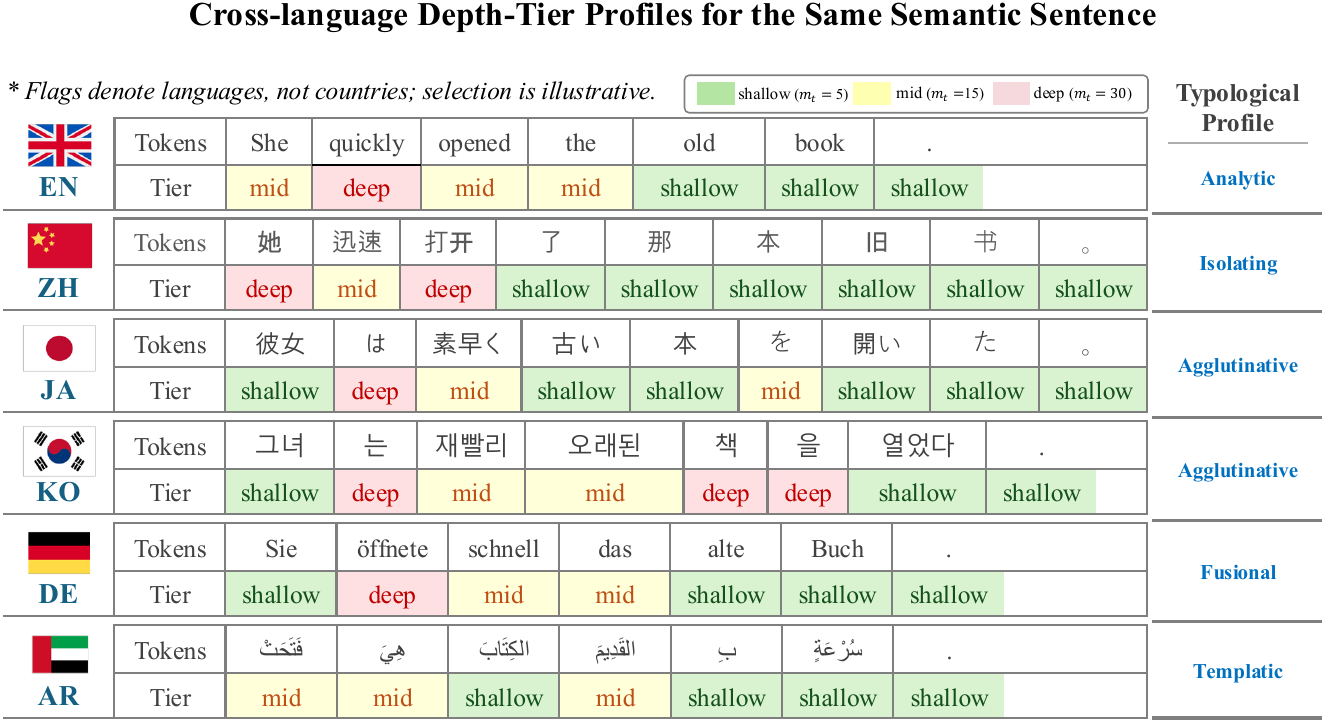}
\caption{
Illustrative cross-language LUNA depth schedules 
for translations of the same semantic sentence.
Each colored cell shows the tournament-depth tier 
selected from normalized next-tag entropy \(\lambda(c_t)\): shallow uses \(m_t=5\), 
mid uses \(m_t=15\), and deep uses \(m_t=30\).
}
\label{fig:cross-language-depth}
\end{figure*}

\section{Method}
\label{sec:method}

\subsection{Linguistic Depth Scheduling}
\label{sec:method-lambda}

LUNA modulates the fixed-depth SynthID-Text backbone 
by choosing the tournament depth from a linguistic signal.
For language \(L\), let \(Q'\) denote the 
next fine-grained part-of-speech tag 
after context \(c\), \(\mathcal{S}_{L,c}\) 
the observed support of next tags in an 
external calibration corpus, 
and \(K_{L,c}=|\mathcal{S}_{L,c}|\).
With empirical probabilities \(\hat{P}_L(q'\mid c)\), define
\begin{equation}
H_L(c)=-\sum_{q'\in\mathcal{S}_{L,c}}\hat{P}_L(q'\mid c)\log_2\hat{P}_L(q'\mid c),
\end{equation}
\begin{equation}
\lambda_L(c)=
\begin{cases}
0, & K_{L,c}\le 1,\\[1mm]
\dfrac{H_L(c)}{\log_2 K_{L,c}}, & K_{L,c}>1.
\end{cases}
\label{eq:lambda}
\end{equation}
Thus \(\lambda_L(c)\in[0,1]\) measures how diffuse 
the observed next-tag distribution is relative 
to its support.
LUNA estimates these tables on 
CulturaX~\citep{nguyen-etal-2024-culturax}, 
separate from evaluation data.
At generation and detection time, 
the lookup procedure backs off from the primary order to 
lower-order contexts and 
returns \(\lambda_{\mathrm{def}}=0.5\) 
when no supported context is available.

LUNA maps \(\lambda(c_t)\) to a three-tier depth schedule,
\begin{equation}
m_t=
\begin{cases}
m_{\min}, & \lambda(c_t)<\tau_1,\\
m_{\mathrm{mid}}, & \tau_1\le \lambda(c_t)<\tau_2,\\
m_{\max}, & \lambda(c_t)\ge \tau_2,
\end{cases}
\label{eq:depth-schedule}
\end{equation}
with the default schedule \((m_{\min},m_{\mathrm{mid}},m_{\max})=(5,15,30)\).
Thresholds \(\tau_1\) and \(\tau_2\) are 
the frequency-weighted 25th and 75th percentiles
of \(\lambda_L\) on the calibration table.
We adopt a three-tier discretization 
as a simple and auditable instantiation of depth 
scheduling: the schedule has only two free thresholds 
that are calibrated from the same corpus used 
for \(\lambda\), and tier identities are easy to 
inspect during error analysis.
Finer discretizations or a continuous mapping 
\(m_t=f(\lambda(c_t))\) are natural extensions.
The schedule is prefix-measurable 
because \(c_t\), \(\lambda(c_t)\), and \(m_t\) 
are all determined before sampling the current token.

Figure~\ref{fig:cross-language-depth} illustrates 
the typological motivation: 
the same semantic content induces different LUNA 
depth schedules across the six evaluation languages.

\subsection{Variable-Depth Generation and Model-Free Detection}
\label{sec:method-detection}

\begin{figure*}[ht!]
\includegraphics[width=\textwidth]{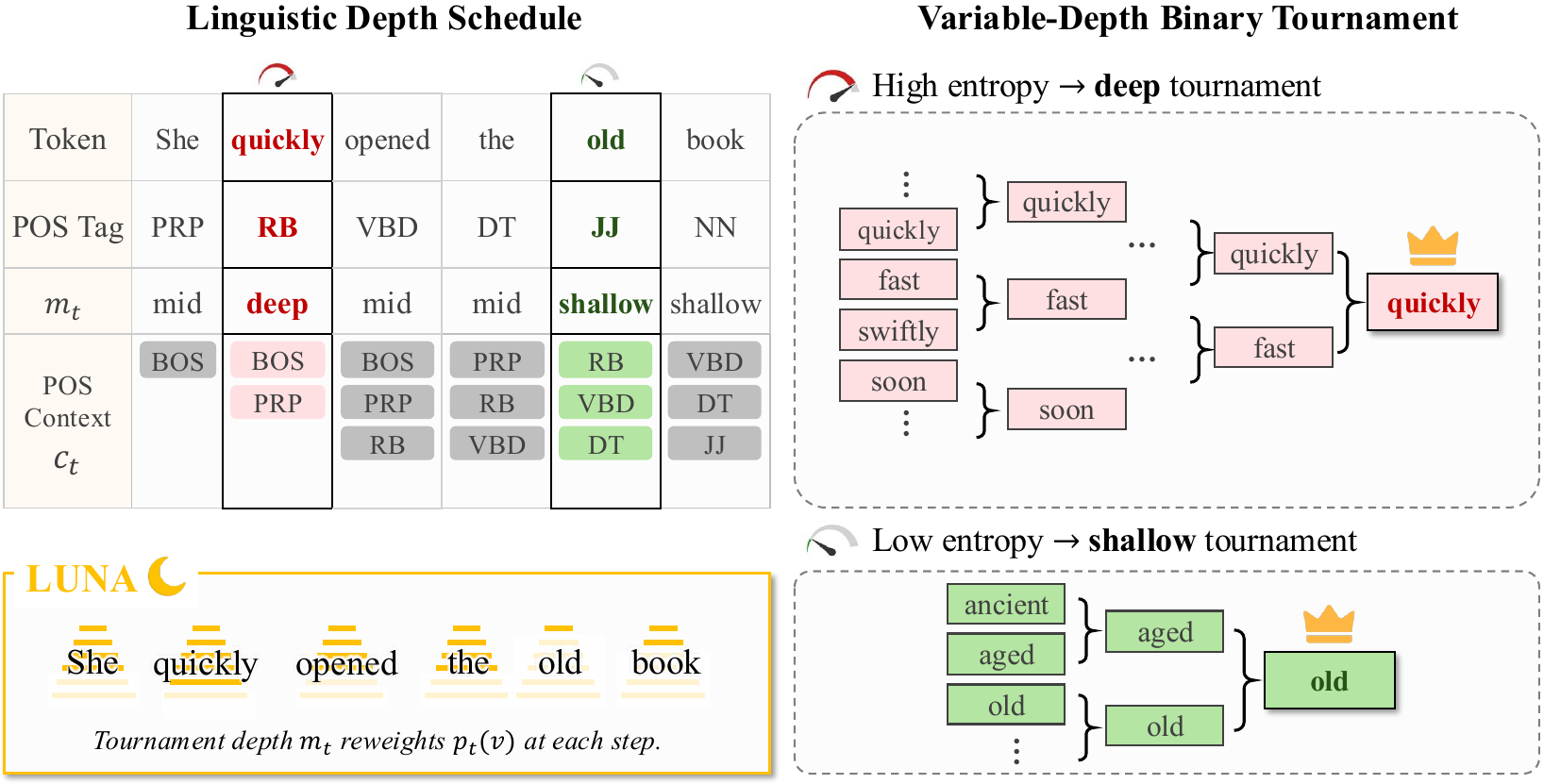}
\caption{
Generation-time operation of LUNA.
For each prefix \(x_{<t}\), 
the base language model supplies the 
next-token distribution \(p_t(v)\), 
while the linguistic branch reconstructs 
the POS context \(c_t\), 
looks up the precomputed normalized next-tag entropy \(\lambda(c_t)\), 
and maps it to a tournament depth \(m_t\).
LUNA then applies an \(m_t\)-layer binary tournament 
that reweights \(p_t(v)\) before sampling \(x_t\).}
\label{fig:luna-overview}
\end{figure*}

LUNA extends the fixed-depth binary tournament 
in Section~\ref{sec:bg-tournament} by replacing 
the constant depth \(m\) with the 
prefix-measurable depth \(m_t\).
Conditioned on a prefix and its depth, 
the current sampling step applies the same 
binary tournament layers used by SynthID-Text.
For notation and implementation, 
we write the binary tournament in its 
probability-rescaling form.
Let \(G_{t,v}^{(\ell)}\in\{0,1\}\) denote 
the value assigned to candidate token \(v\) at 
layer \(\ell\) for position \(t\).
Starting from \(q_t^{(0)}=p_t\), LUNA applies
\begin{equation}
\mu_t^{(\ell)} = \sum_{u\in\mathcal{V}} q_t^{(\ell-1)}(u)G_{t,u}^{(\ell)},
\label{eq:tournament-mu}
\end{equation}
\begin{equation}
q_t^{(\ell)}(v)=q_t^{(\ell-1)}(v)\bigl(1+G_{t,v}^{(\ell)}-\mu_t^{(\ell)}\bigr)
\label{eq:tournament-update}
\end{equation}
for \(\ell=1,\ldots,m_t\), and then samples
\begin{equation}
x_t\sim q_t^{(m_t)}.
\label{eq:generation-sample}
\end{equation}
A repeated-context safeguard leaves the base 
distribution unchanged when the current 
hash context repeats in the recent history; 
the detector skips the same positions.
Figure~\ref{fig:luna-overview} illustrates the 
generation-time operation of LUNA.

Detection uses the text, tokenizer, 
part-of-speech tagger, 
linguistic signal~($\lambda$) table, 
and secret key.
It does not access logits or forward passes of 
the original generation model, 
nor does it run a surrogate model.
The detector aligns tag spans to token positions, 
reconstructs \(c_t\), \(\lambda(c_t)\), and \(m_t\) 
at every valid position, and computes
\begin{equation}
S_t = \sum_{\ell=1}^{m_t}\left(G_{t,x_t}^{(\ell)}-\frac{1}{2}\right),
\label{eq:detection-S}
\end{equation}
\begin{equation}
Z = \frac{\sum_{t\in\mathcal{I}}\omega_t S_t}{\sqrt{\frac{1}{4}\sum_{t\in\mathcal{I}}m_t\omega_t^2}},
\label{eq:detection-z}
\end{equation}
where \(\mathcal{I}\) is the set of valid positions 
and \(\omega_t=\lambda(c_t)\).
Under the random-key null, 
each centered value 
\(G_{t,x_t}^{(\ell)}-1/2\) has variance \(1/4\), 
so the denominator standardizes the weighted sum 
and \(Z\) is comparable to a standard normal score.
Appendix~\ref{app:method-details} gives full 
pseudocode for lookup, generation, and detection.

\subsection{Single-Token Marginal Preservation}
\label{sec:method-theory}

\begin{theorem}[Single-token marginal preservation]
\label{thm:nondistortion}
Fix a prefix \(x_{<t}\) and let \(p_t\) 
be the base distribution passed to the sampler.
Assume that \(m_t=m(x_{<t})\) is prefix-measurable 
and independent of the layer-wise watermark randomness at position \(t\).
Under the standard random-key 
model~\citep{aaronson2022watermark,kuditipudi2024robust,dathathri2024scalable}, 
in which 
\(G_{t,v}^{(\ell)}\overset{\text{iid}}{\sim}\mathrm{Bernoulli}(1/2)\) 
across the index tuples \((t,\ell,v)\), 
the tournament update of 
Equations~\ref{eq:tournament-mu} 
and~\ref{eq:tournament-update} satisfies
\begin{equation*}
\mathbb{E}_{G}\left[\Pr_{\mathrm{LUNA}}(x_t=v\mid x_{<t},G)\right]=p_t(v)
\end{equation*}
for every \(v\in\mathcal{V}\).
\end{theorem}

Theorem~\ref{thm:nondistortion} establishes
a one-step marginal result under the random-key model.
Its role is enabling rather than explanatory:
the theorem certifies that replacing the fixed
tournament depth of the backbone with a
prefix-measurable variable-depth schedule
does not forfeit the random-key guarantee.
It does not claim equality of the
realized fixed-key distribution at a single step,
nor equality of the full joint distribution over sequences.
In particular, because LUNA and the budget-matched
SynthID-Text baseline of
Section~\ref{sec:settings-methods} both satisfy
the theorem, the quality differences reported in
Section~\ref{sec:results} arise from the realized
fixed-key schedules and are established
empirically rather than by the theorem.
The proof follows by conditioning on the prefix
so that \(m_t\) is fixed,
applying the fixed-depth tournament expectation layer
by layer,
and using \(\mathbb{E}[G_{t,v}^{(\ell)}]=1/2\).
Appendix~\ref{app:proof} provides the full proof 
and implementation-level details.

\begin{table}[ht!]
\centering
\small
\setlength{\tabcolsep}{2.5pt}
\begin{tabular}{@{}l l l@{}}
\toprule
\textbf{Lang.} & \textbf{Generation model} & \textbf{POS pipeline} \\
\midrule
EN & Llama-3.2-1B-Instruct & spaCy / PTB \\
ZH & Qwen2.5-0.5B-Instruct & HanLP / CTB \\
KO & EXAONE-3.5-2.4B-Instruct & Kiwi / Sejong \\
JA & Sarashina2.2-3B-Instruct & Sudachi-A / UniDic \\
DE & EuroLLM-1.7B-Instruct & spaCy / STTS \\
AR & Jais-1.3B-Chat & CAMeL Tools / PATB \\
\bottomrule
\end{tabular}
\caption{Evaluation languages, generation models, and part-of-speech 
pipelines.
}
\label{tab:settings-langs}
\end{table}

\section{Experimental Settings}
\label{sec:experimental-settings}

\subsection{Languages and Models}
\label{sec:settings-languages}

The evaluation covers six languages and 
two domains~(Wikipedia, news),
yielding 12 language-by-domain settings.
Each language uses an instruction-tuned generation model 
that natively supports it, 
alongside a 
language-specific part-of-speech~(POS) pipeline.
Table~\ref{tab:settings-langs} summarizes 
the main experimental setup.
Appendix~\ref{app:exp-details} gives 
full model identifiers, part-of-speech backends, 
tagsets, and selected context orders.
For perplexity-based quality evaluation, 
we use Qwen2.5-1.5B~\citep{yang2024qwen2} 
as a shared reference model across languages.

\subsection{Datasets}
\label{sec:settings-data}

LUNA estimates $\lambda$ tables 
from CulturaX~\citep{nguyen-etal-2024-culturax}, 
using 20{,}000 held-out records per language 
with a length filter of 300 to 4000 characters.
The same held-out corpus supplies calibration for STELA.
No evaluation prompt or generated output enters 
the calibration corpus.
We use two dataset families:
Wikipedia continuations for all six languages~\citep{wikimedia2023wikipedia}, 
and news continuations from 
XL-Sum~\citep{hasan-etal-2021-xl} 
for English, Chinese, Korean, Japanese, and Arabic, 
plus MLSum~\citep{scialom-etal-2020-mlsum} for German.
Each language-by-domain setting contains 500 records,
so each algorithm runs on 6{,}000 evaluation records.

\subsection{Baselines and Generation Protocol}
\label{sec:settings-methods}

We compare LUNA with eight baselines: 
KGW, EWD, SWEET, MorphMark, 
STELA, GumbelSoft, EXP, and SynthID-Text.
SynthID-Text is configured to match the 
expected tournament budget 
\(B=\mathbb{E}[2^{m_t}]\) 
induced 
by the LUNA depth ladder, 
equalizing the expected per-token tournament budget across the two methods;
the matching formula appears in Appendix~\ref{app:synthid-budget}.
At the nominal tier proportions, the matched budget is
\(B_0=268{,}451{,}848\approx 2^{28}\),
which realizes a nearly constant depth-$28$ schedule;
the match equalizes expected candidate counts
rather than realized distortion~(Appendix~\ref{app:synthid-budget}).
All methods sample with temperature $0.7$, 
nucleus probability $0.95$, 
no top-$k$ cap, 
and $200$--$256$ new tokens; 
Qwen2.5-0.5B uses repetition 
penalty $1.1$, others $1.0$.
Watermarked, unwatermarked, and human-reference texts 
are truncated to at 
most $256$ generation-tokenizer tokens 
before detection so that model-aware 
detectors fit within GPU memory at equal evidence length.
Detailed seeds, context orders, 
and method-specific 
hyperparameters appear in 
Appendix~\ref{app:watermark-hparams}; 
experiments run on a single NVIDIA RTX 3090 
GPU with $24$~GB of memory.

\begin{table*}[ht!]
\centering
\small
\setlength{\tabcolsep}{4.5pt}
\begin{tabular}{@{}l c c c c c c c c@{}}
\toprule
& \multicolumn{2}{c}{\textbf{Detection}} & \multicolumn{6}{c}{\textbf{Quality Preservation}} \\
\cmidrule(lr){2-3}\cmidrule(lr){4-9}
\textbf{Method} & AUROC & TPR@5\% 
& $|\Delta\mathrm{PPL}_{\mathrm{med}}|$ & 95\% CI
& $|\Delta\textsc{SBleu}|$ & $|\Delta\mathrm{Dist}_1|$ 
& $|\Delta\mathrm{Surp}|$ & $|\Delta\mathrm{Entr}|$ \\
\midrule
KGW          & 0.9982          & 0.9952          & 1.290 & [0.625, 2.133] & 0.0063 & 0.0106 & 0.1357 & 0.0789 \\
EWD          & \textbf{0.9990} & \textbf{0.9972} & 1.115 & [0.554, 1.820] & 0.0063 & 0.0101 & 0.1186 & 0.0645 \\
SWEET        & 0.9985          & 0.9950          & 0.915 & [0.474, 1.482] & 0.0050 & 0.0068 & 0.0989 & 0.0589 \\
MorphMark    & 0.9902          & 0.9643          & 0.425 & [0.158, 0.734] & 0.0024 & 0.0052 & 0.0442 & 0.0275 \\
STELA        & 0.9982          & 0.9953          & 1.182 & [0.620, 1.911] & 0.0065 & 0.0114 & 0.1250 & 0.0711 \\
GumbelSoft   & 0.9899          & 0.9778          & 1.202 & [0.485, 2.112] & 0.0575 & 0.0653 & 0.1473 & 0.1370 \\
EXP          & 0.9876          & 0.9777          & 1.310 & [0.509, 2.286] & 0.0711 & 0.0728 & 0.1659 & 0.1640 \\
SynthID-Text & 0.9972          & 0.9928          & 0.463 & [0.219, 0.751] & 0.0040 & 0.0062 & 0.0514 & 0.0396 \\
\midrule
\textbf{LUNA} & 0.9959 & 0.9868 
& \textbf{0.045} & \textbf{[0.022, 0.073]} 
& \textbf{0.0016} & \textbf{0.0029} & \textbf{0.0054} & \textbf{0.0116} \\
\bottomrule
\end{tabular}
\caption{
Main detection and quality preservation results, 
$12$-setting mean.
SBleu, $\mathrm{Dist}_1$, Surp, and Entr abbreviate 
Self-BLEU, 
Distinct-$1$, surprisal, and entropy.
}
\label{tab:main-results}
\end{table*}

\subsection{Evaluation Metrics}
\label{sec:settings-metrics}

\paragraph{Detection metrics.}
We use AUROC and TPR at $5\%$ FPR.
Both compare watermarked outputs with unwatermarked outputs 
generated by the same base model 
from the same prompts.
AUROC summarizes the full ROC curve;
TPR at $5\%$ FPR fixes
a deployment-relevant operating point.
Detection against human-written references is
reported in Section~\ref{sec:results-human}
and Appendix~\ref{app:human-negatives}.

\paragraph{Quality metrics.}
For each text-level quality statistic $Q$, 
we form the absolute setting-level shift 
$|\Delta Q|=|Q_{\mathrm{w}}-Q_{\mathrm{u}}|$, 
where $Q_{\mathrm{w}}$ is computed on the watermarked outputs of a setting and 
$Q_{\mathrm{u}}$ on the unwatermarked outputs of the same setting and prompts.
We define five statistics covering complementary notions of distortion.
$|\Delta\mathrm{PPL}_{\mathrm{med}}|$ uses median perplexity under 
Qwen2.5-1.5B and 
captures the likelihood of the generated text under the reference model.
$|\Delta\textsc{Self-BLEU}|$ measures corpus-level Self-BLEU for 
intra-output lexical repetition.
$|\Delta\mathrm{Distinct\text{-}1}|$ computes the Distinct-1 ratio for 
unigram diversity at the surface level.
$|\Delta\mathrm{Surprisal}|$ and $|\Delta\mathrm{Entropy}|$
evaluate the mean token-level surprisal
and predictive entropy under the same
reference model,
capturing distortion at the next-token-distribution level.
These five statistics form two related diagnostic
families rather than five independent tests:
the perplexity, surprisal, and entropy shifts probe
the reference-model distribution,
while the Self-BLEU and Distinct-$1$ shifts probe
lexical diversity.
We therefore read agreement across the two families,
rather than the metric count,
as the meaningful evidence of preservation.

\paragraph{Aggregation and confidence intervals.}
All statistics are aggregated at the setting level: 
we first compute each 
statistic within each of the $12$ 
language-by-domain settings and then report 
the mean over settings.
Bootstrap $95\%$ confidence intervals resample the 
$12$ settings with 
replacement over $1000$ iterations.
Section~\ref{sec:results-main} 
reports both the mean and the bootstrap 
interval for $|\Delta\mathrm{PPL}_{\mathrm{med}}|$; 
intervals for the other four quality metrics appear in 
Appendix~\ref{app:bootstrap-cis}.

\section{Experimental Results}
\label{sec:results}

\subsection{Main Detection-Quality Results}
\label{sec:results-main}

Table~\ref{tab:main-results} reports the 
experimental results.
For every method that exposes 
a part-of-speech context order as a 
hyperparameter, namely LUNA and STELA,
results use the per-algorithm per-setting 
best context order from 
Table~\ref{tab:selected-k}~(Appendix~\ref{app:settings-details}).
Bold values mark the best entry per column.

\paragraph{Detection saturation.}
Six methods achieve AUROC above $0.995$: 
EWD, SWEET, KGW, STELA, SynthID-Text, and LUNA.
Within this regime, the AUROC gap between EWD and LUNA is only $0.0031$, while the TPR-at-$5\%$-FPR gap is $0.0104$.
Both gaps are small in absolute terms; 
we therefore focus primarily on the detection–quality 
trade-off rather than the ranking within 
this near-saturated regime.
Furthermore, EWD and SWEET require language-model 
forward passes at detection time, 
while KGW, STELA, SynthID-Text, and LUNA detect from 
text, tokenizer, tagger, and 
secret key alone; 
LUNA therefore matches the strongest 
model-based detector within these 
margins without requiring the language model 
at verification.

\begin{table*}[ht!]
\centering
\small
\setlength{\tabcolsep}{4.5pt}
\begin{tabular}{@{}l c c c c c c c@{}}
\toprule
& \multicolumn{2}{c}{\textbf{Detection}} & \multicolumn{5}{c}{\textbf{Quality Preservation Factor (Control / LUNA)}} \\
\cmidrule(lr){2-3}\cmidrule(lr){4-8}
\textbf{Comparison} 
& $\Delta$AUROC & $\Delta$TPR@5\% 
& $|\Delta\mathrm{PPL}_{\mathrm{med}}|$ 
& $|\Delta\textsc{SBleu}|$ & $|\Delta\mathrm{Dist}_1|$ 
& $|\Delta\mathrm{Surp}|$ & $|\Delta\mathrm{Entr}|$ \\
\midrule
LUNA $-$ STELA                        & $-0.0023$ & $-0.0085$ & $26.41\times$ & $4.07\times$ & $3.96\times$ & $22.99\times$ & $6.12\times$ \\
LUNA $-$ SynthID-Text                 & $-0.0013$ & $-0.0060$ & $10.35\times$ & $2.53\times$ & $2.15\times$ & $9.45\times$  & $3.41\times$ \\
LUNA $-$ SynthID-Text-Entropy         & $-0.0001$ & $-0.0007$ & $1.76\times$  & $1.59\times$ & $0.92\times$ & $1.69\times$  & $1.70\times$ \\
\bottomrule
\end{tabular}
\caption{Controlled comparisons against LUNA, averaged over the \(12\) settings.
Detection columns report LUNA minus the control; quality columns report the
control divided by LUNA, so factors above \(1\) indicate that LUNA changes the
metric less.}
\label{tab:ablations}
\end{table*}

\paragraph{Dominant multi-metric quality preservation.}
LUNA records the lowest mean shift on every one of the five quality metrics.
Relative to the closest baseline~(MorphMark across all five metrics), 
LUNA achieves a $9.5\times$ reduction on
$|\Delta\mathrm{PPL}_{\mathrm{med}}|$, 
$1.5\times$ reduction on $|\Delta\textsc{Self-BLEU}|$, 
$1.8\times$ on 
$|\Delta\mathrm{Distinct\text{-}1}|$, 
$8.1\times$ on $|\Delta\mathrm{Surprisal}|$, 
and $2.4\times$ on $|\Delta\mathrm{Entropy}|$.
The dominance covers complementary aspects of distortion at once: 
the language-model probability of the 
generated text~($|\Delta\mathrm{PPL}_{\mathrm{med}}|$), 
its lexical structure~($|\Delta\textsc{Self-BLEU}|$, $|\Delta\mathrm{Distinct\text{-}1}|$), 
and the realized next-token-distribution 
statistics~($|\Delta\mathrm{Surprisal}|$, $|\Delta\mathrm{Entropy}|$).

\paragraph{Bootstrap separation on the quality metric.}
The bootstrap intervals summarize between-setting
variation under the disclosed selection protocol;
they do not account for within-setting
context-order selection~(Appendix~\ref{app:settings-details}).
Under this scope, the LUNA interval $[0.022, 0.073]$
does not overlap any baseline
interval,
and the next-lowest baseline lower bound is
$0.158$~(MorphMark).
Appendix~\ref{app:bootstrap-cis} reports the
intervals for the remaining quality metrics.

\paragraph{Fixed-order robustness.}
The quality advantage does not depend on
per-setting context-order selection.
Holding the order fixed at $k\in\{2,3,4\}$ for all
twelve settings, LUNA attains AUROC between
$0.9947$ and $0.9958$, TPR at $5\%$ FPR between
$0.9818$ and $0.9862$, and
$|\Delta\mathrm{PPL}_{\mathrm{med}}|$ between
$0.0999$ and $0.1376$~(Appendix~\ref{app:k-analysis}).
At $k=2$ and $k=3$, LUNA records smaller shifts
than every baseline on all five quality metrics,
and at $k=4$ on four of five;
even at LUNA's largest fixed-order perplexity
shift, the closest baseline shift is $3.09$ times
as large~($0.425$ for MorphMark against $0.1376$).
Context-order selection therefore changes the
exact headline values, not the substantive
conclusion.

\subsection{Detection Against Human-Written References}
\label{sec:results-human}

The main protocol scores watermarked outputs
against unwatermarked outputs of the same model
because this matched-source comparison isolates
watermark-specific evidence:
the two classes share the generator, prompts,
decoding configuration, domains, and length
handling, and differ only in whether the keyed
watermark is applied.
Deployment, however, also asks whether genuine
human writing is falsely flagged.
We therefore repeat detection with the gold human
continuation of each record as the negative class,
over all twelve settings and $500$ records per
setting, under the same context-order rule as
Table~\ref{tab:main-results}.
LUNA attains mean AUROC $0.9960$ and mean TPR
$0.9870$ when the $5\%$ FPR operating point is
calibrated on the human negatives,
closely matching the main-protocol values of
$0.9959$ and $0.9868$;
across the nine methods, the two protocols agree
within $0.003$ AUROC.
Appendix~\ref{app:human-negatives} reports the
full per-setting tables and analysis.

Two clarifications delimit this claim.
First, the schedule directs capacity toward
high-$\lambda$ contexts, whose continuations are
less predictable, and
Equation~\ref{eq:detection-z} down-weights
low-$\lambda$ positions,
so syntactically predictable formal writing
receives neither greater insertion depth nor
greater detection weight.
Second, these results describe the evaluated
references; they do not imply a universal
fixed-threshold false-positive rate for arbitrary
human corpora, which a deployment would calibrate
separately.

\subsection{Ablation Study}
\label{sec:results-ablations}

Table~\ref{tab:ablations} compares LUNA with three targeted references that
isolate the main design decisions behind the method.
STELA is the closest linguistic baseline: 
it uses a corpus-estimated
part-of-speech signal, 
yet injects that signal through a distortionary
green-list bias.
SynthID-Text is the closest tournament baseline: 
it uses a non-distortionary
binary tournament backbone, 
yet allocates watermark capacity without a
linguistic signal.
SynthID-Text-Entropy is a controlled baseline 
introduced in this paper.
It replaces the corpus-estimated linguistic signal 
of LUNA with
language-model entropy, 
thereby testing whether model-side uncertainty can
substitute for the proposed POS-context signal.
Appendix~\ref{app:synthid-entropy} gives 
the full construction.

\paragraph{Linguistic signal without non-distortion: STELA.}
STELA and LUNA both use corpus-estimated 
POS-context uncertainty.
The difference lies in the sampling backbone: 
STELA injects the signal through
green-list logit bias, 
whereas LUNA uses it to modulate the depth of a
non-distortionary tournament sampler.
This comparison shows the value of 
replacing a distortionary linguistic
watermark with a non-distortionary tournament mechanism.
At comparable detection~(AUROC and TPR@5\% within 
$0.0023$ and $0.0085$ respectively), 
LUNA reduces the five quality shifts by
\(3.96\times\) to \(26.41\times\).

\paragraph{Tournament sampling without linguistic scheduling: SynthID-Text.}
SynthID-Text and LUNA share the binary tournament backbone.
The difference is the source of the schedule: 
SynthID-Text uses
prefix-hash randomness, 
while LUNA uses \(\lambda(c_t)\) to place more
capacity in high-uncertainty POS contexts.
This comparison isolates the effect of
linguistic scheduling within the same
tournament family.
LUNA reduces all five quality shifts by
\(2.15\times\) to \(10.35\times\)
while retaining nearly the same AUROC and TPR@5\%.
Because the budget match equalizes expected
candidate counts rather than realized
distortion~(Appendix~\ref{app:synthid-budget}),
it drives SynthID-Text to a nearly constant
depth-$28$ schedule;
the \(10.35\times\) factor is therefore specific
to this resource-matching protocol,
and the model-entropy control below provides the
more direct adaptive comparison.

\paragraph{Model entropy instead of linguistic entropy: SynthID-Text-Entropy.}
SynthID-Text-Entropy is a new controlled baseline 
designed for this study.
It asks whether model-derived entropy 
can replace the external linguistic
signal used by LUNA.
The variant keeps the SynthID-Text tournament family 
and budget matching, yet
uses language-model entropy as the adaptive signal 
rather than the
corpus-estimated POS-context entropy used by LUNA.
This gives a strong model-aware comparison point: detection is nearly
identical to LUNA, 
with gaps of only \(-0.0001\) AUROC and 
\(-0.0007\)
TPR@5\%.
The detector requires language-model forward passes 
at verification time,
which sacrifices model-free detection, 
and LUNA still improves four of five
quality metrics by \(1.59\times\) to \(1.76\times\) 
on average.

\section{Conclusion}
\label{sec:conclusion}

As LLM-generated text becomes difficult to 
distinguish from human writing, provenance 
signals are now a practical requirement for 
platforms, publishers, and regulators; 
a deployable watermark should preserve quality, 
and verify without model access. 
LUNA combines part-of-speech context entropy 
with a non-distortionary 
tournament sampler to jointly satisfy 
single-token non-distortion, 
model-free detection, and linguistic adaptivity.
Across six typologically diverse languages 
and two domains, 
it records the lowest mean shift across 
the five quality metrics 
and is the only method reaching AUROC $>0.99$ 
and 
$|\Delta\mathrm{PPL}_{\mathrm{med}}|<0.1$ 
in a majority of settings.

The central take-away is that the adaptive 
signal of a watermark need not come from the 
model it marks: 
corpus-estimated syntactic uncertainty 
allocates capacity well while remaining 
external to the generator. 
This turns adaptivity into an inspectable 
artifact, a $\lambda$ table that can be 
published, audited, and versioned, in line 
with provenance regulation. 

One mechanism holds from analytic English to 
templatic Arabic; 
the natural next step is low-resource 
languages, where taggers are weaker and 
calibration corpora smaller. 
Qualitative analysis of watermarked text 
would complement our quantitative evidence. 

\section*{Limitations}
\label{sec:limitations}

LUNA uses part-of-speech context entropy as a 
linguistic proxy for watermark capacity. 
This proxy captures syntactic uncertainty 
rather than every form of linguistic choice. 
It does not directly model semantic alternatives, 
discourse structure, pragmatic constraints, or register. 
The empirical results suggest that 
syntactic uncertainty provides a useful control signal, 
while richer linguistic schedules could combine 
POS context with morphology, dependency structure, 
discourse state, or semantic classes. 
LUNA also discretizes \(\lambda(c_t)\) into 
three depth tiers; finer-grained tiers or a 
continuous mapping \(m_t=f(\lambda(c_t))\) 
are natural extensions that we leave to future work. 
Such extensions would test how much of the watermark 
capacity arises from syntax alone and how much 
comes from broader linguistic organization.

The method also depends on language-specific analyzers 
and entropy tables. 
We use deterministic POS pipelines and keep 
the same tagger and tagset across calibration, 
generation, and detection. 
This design makes the schedule auditable, 
yet it transfers responsibility to the 
linguistic preprocessing layer. 
Languages with limited taggers, unstable segmentation, 
code switching, or domain-specific orthography 
may require additional calibration. 
Future work can study tagger uncertainty, 
multilingual tagset normalization, 
and analyzer ensembles that preserve 
model-free detection while reducing dependence 
on a single preprocessing pipeline.

The theoretical guarantee has a precise scope. 
LUNA preserves single-token marginals under the 
standard random-key model for the non-distortionary 
tournament sampler. 
This statement does not imply equality of the full 
joint sequence distribution for a fixed key, 
and it does not provide an inherent guarantee 
against paraphrase, translation, editing, or 
adversarial attacks. 
These transformations can change the observed 
POS sequence, the reconstructed schedule, 
or the keyed evidence. 
Our evaluation therefore treats robustness 
as an empirical question rather than as a 
theorem-level property.

Finally, model-free detection does not mean 
infrastructure-free detection. 
A verifier still needs the tokenizer, the POS analyzer, 
the entropy table, and the secret key. 
This requirement is substantially weaker than access 
to target-model logits or surrogate forward passes, 
and it supports public-verification scenarios 
more naturally than model-dependent adaptive schemes. 
Nevertheless, deployment would need key management, 
versioning of entropy tables, 
and documented analyzer configurations. 
These operational requirements define a 
concrete path for extending LUNA from a 
research watermark to an auditable multilingual 
provenance system.

\section*{Acknowledgments}
This research was supported by the NRF grant~(RS-2025-00562134) and the AI Graduate School Program~(RS-2020-II201361) funded by the Korean government.

\bibliography{custom}

\appendix

\section{Method Details}
\label{app:method-details}

This appendix provides the implementation details 
omitted from the main text for space.
Algorithm~\ref{alg:lookup} gives the 
deterministic entropy lookup with order backoff.
Algorithm~\ref{alg:gen} gives per-position generation, 
and Algorithm~\ref{alg:det} gives model-free detection.

\begin{algorithm}[ht!]
\caption{Order-backoff lookup of normalized next-tag entropy.}
\label{alg:lookup}
\begin{algorithmic}[1]
\Require POS context $c_t$; language $L$; lookup tables and thresholds.
\Ensure Normalized next-tag entropy $\lambda\in[0,1]$.
\For{$r=k_{\mathrm{primary}},k_{\mathrm{primary}}-1,\ldots,2$}
  \State $c^{(r)}\gets$ truncate $c_t$ to the last $r-1$ tags
  \If{$c^{(r)}\in\mathcal{T}_L^{(r)}$ \textbf{and} $N_L(c^{(r)})\ge\nu$}
    \State \Return $\lambda_L(c^{(r)})$
  \EndIf
\EndFor
\State \Return $\lambda_{\mathrm{def}}$
\end{algorithmic}
\end{algorithm}

\subsection{Entropy Lookup with Order Backoff}
\label{app:lookup}

We summarize the additional notation used by 
Algorithm~\ref{alg:lookup}. 
For language $L$ and order $r\in\{2,\ldots,k_{\mathrm{primary}}\}$, 
let $\mathcal{T}_L^{(r)}$ denote the set of 
length-$(r{-}1)$ POS contexts observed in the 
calibration corpus, 
$N_L(c)$ denote the empirical occurrence count 
of context $c$ in that corpus, 
and $\nu$ denote a fixed minimum-count threshold 
that controls when a stored value is reused. 
The threshold $\nu$ is shared across orders 
and languages, and is chosen on the calibration 
corpus so that stored \(\lambda\) values rely only 
on contexts with stable empirical estimates.

The lookup starts at the primary order 
$k_{\mathrm{primary}}$ and backs off through 
lower orders down to order $2$.
It returns a stored value only when the context exists 
in $\mathcal{T}_L^{(r)}$ 
and its empirical frequency reaches the threshold $\nu$.
If no supported context appears, 
it returns $\lambda_{\mathrm{def}}=0.5$.

\begin{algorithm}[ht!]
\caption{LUNA generation at position $t$.}
\label{alg:gen}
\begin{algorithmic}[1]
\Require Prefix $x_{<t}$; distribution $p_t$; keys; $\Phi$; schedule; tagger $\mathcal{A}$; history $\mathcal{H}$.
\Ensure Next token $x_t$.
\State $c_t\gets\mathrm{POSContext}(\mathcal{A},x_{<t})$
\State $\lambda\gets\mathrm{Lookup}(c_t)$
\State $m_t\gets\mathrm{MapToDepth}(\lambda)$ using Equation~\ref{eq:depth-schedule}
\State $h\gets\mathrm{HashContext}(x_{<t})$
\State $r\gets\mathbf{1}\{h\in\mathcal{H}\}$
\State $\mathcal{H}\gets\mathrm{UpdateHistory}(\mathcal{H},h)$
\If{$r=1$}
  \State \Return $x_t\sim p_t$
\EndIf
\State $q^{(0)}\gets p_t$
\For{$\ell=1$ \textbf{to} $m_t$}
  \State $G_v^{(\ell)}\gets\Phi(k_\ell,h,v)$ for each $v\in\mathcal{V}$
  \State $\mu^{(\ell)}\gets\sum_{u\in\mathcal{V}}q^{(\ell-1)}(u)G_u^{(\ell)}$
  \State $q^{(\ell)}(v)\gets q^{(\ell-1)}(v)(1+G_v^{(\ell)}-\mu^{(\ell)})$
\EndFor
\State \Return $x_t\sim q^{(m_t)}$
\end{algorithmic}
\end{algorithm}

\subsection{Generation and Detection Algorithms}
\label{app:algorithms}

Algorithm~\ref{alg:gen} specifies per-position
generation with the repeated-context safeguard,
and Algorithm~\ref{alg:det} specifies model-free
detection;
both consume the entropy lookup of
Algorithm~\ref{alg:lookup} and the depth schedule
of Equation~\ref{eq:depth-schedule}.

\subsection{Proof of Theorem~\ref{thm:nondistortion}}
\label{app:proof}

\begin{algorithm}[ht!]
\caption{LUNA model-free detection.}
\label{alg:det}
\begin{algorithmic}[1]
\Require Text $x=(x_1,\ldots,x_T)$; tokenizer; tagger $\mathcal{A}$; $\lambda$ table; keys; threshold $\gamma$.
\Ensure Decision: watermarked or not.
\State Run $\mathcal{A}$ on the decoded full text and align tag spans to token positions
\State $\mathcal{I}\gets\emptyset$; $\mathcal{H}\gets\emptyset$
\For{$t=1$ \textbf{to} $T$}
  \State $h\gets\mathrm{HashContext}(x_{<t})$
  \State $r\gets\mathbf{1}\{h\in\mathcal{H}\}$
  \State $\mathcal{H}\gets\mathrm{UpdateHistory}(\mathcal{H},h)$
  \If{$x_t$ is EOS \textbf{or} $r=1$}
    \State \textbf{continue}
  \EndIf
  \State Recover $c_t$ before position $t$
  \State $\lambda\gets\mathrm{Lookup}(c_t)$
  \State $m_t\gets\mathrm{MapToDepth}(\lambda)$
  \State $\omega_t\gets\lambda$
  \State $S_t\gets\sum_{\ell=1}^{m_t}(G_{t,x_t}^{(\ell)}-1/2)$
  \State $\mathcal{I}\gets\mathcal{I}\cup\{t\}$
\EndFor
\State Compute $Z$ with Equation~\ref{eq:detection-z}
\State \Return $\mathbf{1}\{Z>\gamma\}$
\end{algorithmic}
\end{algorithm}

Condition on the prefix $x_{<t}$.
The POS reconstruction returns $c_t$, 
the lookup returns $\lambda(c_t)$, 
and the schedule fixes $m_t$ before token $x_t$ is sampled.
The current step therefore reduces to 
fixed-depth binary tournament sampling with depth $m_t$.
Let $q^{(0)}=p_t$.
For layer $\ell$, condition on previous layers, 
so $q^{(\ell-1)}$ is fixed.
Under the random-key model, 
the binary values $G_v^{(\ell)}$ 
are independent of $q^{(\ell-1)}$ 
and satisfy $\mathbb{E}[G_v^{(\ell)}]=1/2$, so
\begin{align*}
\mathbb{E}_{G^{(\ell)}}[q^{(\ell)}(v)\mid q^{(\ell-1)}]
&=q^{(\ell-1)}(v)\left(1+\frac{1}{2}-\frac{1}{2}\right)\\
&=q^{(\ell-1)}(v).
\end{align*}
Iterating across the active layers and applying the 
tower property of conditional expectation yields 
$\mathbb{E}_G[q^{(m_t)}(v)]=p_t(v)$.
Since $x_t$ is drawn from $q^{(m_t)}$ conditional on $G$, 
$\Pr_{\mathrm{LUNA}}(x_t=v\mid x_{<t},G)=q^{(m_t)}(v)$, 
and taking expectation over $G$ gives 
$\mathbb{E}_G[\Pr_{\mathrm{LUNA}}(x_t=v\mid x_{<t},G)]=p_t(v)$.
Prefix measurability ensures that $m_t$ does not 
depend on the current sampled token, 
so it remains fixed throughout this argument.

\begin{table}[H]
\centering
\small
\setlength{\tabcolsep}{4pt}
\begin{tabular}{@{}l l@{}}
\toprule
\textbf{Lang.} & \textbf{POS backend and tagset} \\
\midrule
EN & spaCy \texttt{en\_core\_web\_md} / PTB \\
ZH & HanLP CTB9 POS + coarse tokenizer / CTB \\
KO & Kiwi / Sejong \\
JA & SudachiPy SplitMode A / UniDic \\
DE & spaCy \texttt{de\_core\_news\_md} / STTS \\
AR & CAMeL Tools default POS tagger / PATB \\
\bottomrule
\end{tabular}
\caption{POS backends and tagsets used by LUNA. 
These choices match the tagsets used to build the corresponding $\lambda$ tables.}
\label{tab:taggers}
\end{table}

\section{Experimental Setting Details}
\label{app:exp-details}

\subsection{Language Typology, Models, and POS Pipelines}
\label{app:models-taggers}

Table~\ref{tab:taggers} lists the
POS backends and tagsets used at entropy estimation,
generation, and detection time.
The backends are spaCy~\citep{spacy2020} for
English and German,
HanLP~\citep{he-choi-2021-stem} for Chinese,
Kiwi~\citep{lee2024kiwi} for Korean,
SudachiPy~\citep{takaoka-etal-2018-sudachi} for
Japanese, and
CAMeL Tools~\citep{obeid-etal-2020-camel} for
Arabic.
For every language, the same tagger and tagset
are used across these three stages.

Table~\ref{tab:model-ids} lists the full 
generation-model identifiers used in the experiments.

\begin{table}[H]
\centering
\footnotesize
\setlength{\tabcolsep}{4pt}
\begin{tabular}{@{}l p{0.72\columnwidth}@{}}
\toprule
\textbf{Lang.} & \textbf{Model identifier} \\
\midrule
EN & \texttt{meta-llama/Llama-3.2-1B-Instruct} \\
ZH & \texttt{Qwen/Qwen2.5-0.5B-Instruct} \\
KO & \begin{tabular}[t]{@{}l@{}}\texttt{LGAI-EXAONE/}\\\texttt{EXAONE-3.5-2.4B-Instruct}\end{tabular} \\
JA & \begin{tabular}[t]{@{}l@{}}\texttt{sbintuitions/}\\\texttt{sarashina2.2-3b-instruct-v0.1}\end{tabular} \\
DE & \texttt{utter-project/EuroLLM-1.7B-Instruct} \\
AR & \texttt{inceptionai/jais-family-1p3b-chat} \\
\bottomrule
\end{tabular}
\caption{Generation-model identifiers.}
\label{tab:model-ids}
\end{table}

\subsection{Budget Matching for SynthID-Text}
\label{app:synthid-budget}

The SynthID-Text baseline 
uses the same binary tournament update as LUNA 
and matches the expected tournament budget induced 
by the LUNA depth ladder.
This budget matching removes the linguistic signal 
from the comparison: 
the depth is derived from a prefix hash and 
a salt rather than from $\lambda(c_t)$, 
and detection uses uniform weights $\omega_t=1$.
The schedule chooses between adjacent depths 
$m_{\mathrm{floor}}=\lfloor\log_2 B\rfloor$ 
and $m_{\mathrm{ceil}}=\lceil\log_2 B\rceil$ 
so that
\[
\mathbb{E}[2^{m_t}]=(1-p_{\mathrm{ceil}})2^{m_{\mathrm{floor}}}+p_{\mathrm{ceil}}2^{m_{\mathrm{ceil}}}=B.
\]
If $m_{\mathrm{floor}}=m_{\mathrm{ceil}}$, 
the schedule uses that depth deterministically.
Otherwise,
\begin{equation}
p_{\mathrm{ceil}}=\frac{B-2^{m_{\mathrm{floor}}}}{2^{m_{\mathrm{ceil}}}-2^{m_{\mathrm{floor}}}}.
\end{equation}
When calibration supplies language-specific 
tier proportions $(p_{\mathrm{low}},p_{\mathrm{mid}},p_{\mathrm{high}})$,
the matched budget is
\begin{equation}
B=p_{\mathrm{low}}2^{m_{\min}}+p_{\mathrm{mid}}2^{m_{\mathrm{mid}}}+p_{\mathrm{high}}2^{m_{\max}}.
\end{equation}
At nominal proportions $(0.25,0.5,0.25)$ and
ladder $(5,15,30)$, this formula gives $B_0=268{,}451{,}848$.
Since $2^{28}=268{,}435{,}456$, the matched budget
satisfies $B_0\approx 2^{28}$, and the two-point
schedule selects $m_{\mathrm{floor}}=28$ with
probability $1-p_{\mathrm{ceil}}\approx 0.99994$:
budget-matched SynthID-Text runs a nearly
constant depth-$28$ tournament.
We emphasize the scope of this construction.
It matches the expected number of tournament
candidates, not realized distortion, so it
defines a resource-matched rather than
distortion-equivalent operating point,
and quality ratios measured against it,
including the $10.35\times$ perplexity-shift
factor in Table~\ref{tab:ablations},
are specific to this protocol.

\begin{table*}[!ht]
\centering
\small
\setlength{\tabcolsep}{4pt}
\begin{tabular}{@{}l p{0.78\textwidth}@{}}
\toprule
\textbf{Method} & \textbf{Main settings} \\
\midrule
KGW & $\gamma=0.5$, $\delta=2.0$ \\
EWD & KGW generation with entropy-weighted detection; $\gamma=0.5$, $\delta=2.0$ \\
SWEET & Entropy-thresholded insertion and detection; $\gamma=0.5$, $\delta=2.0$ \\
MorphMark & Adaptive insertion strength; $\gamma=0.5$, $\delta=1.3$ \\
STELA & POS-conditioned green-list watermark; $\gamma=0.5$, $\delta=2.0/\mathbb{E}[\lambda(c_t)]$ per language \\
EXP & MarkLLM implementation \\
GumbelSoft & Official implementation of GumbelSoft \\
SynthID-Text & MarkLLM implementation; LUNA-matched expected tournament budget \\
LUNA & Depth ladder $(5,15,30)$, history-window size $|\mathcal{H}|=30$, $\lambda_{\mathrm{def}}=0.5$, frequency-weighted $25$th and $75$th percentile depth thresholds \\
\bottomrule
\end{tabular}
\caption{Watermark-specific settings used in the primary comparison.}
\label{tab:watermark-hparams}
\end{table*}

\subsection{Watermark Baselines and Hyperparameters}
\label{app:watermark-hparams}

Table~\ref{tab:watermark-hparams} 
summarizes the main watermark-specific settings.
The KGW-family baselines follow the 
MarkLLM~\citep{pan-etal-2024-markllm} implementations used in the experiments.
The SynthID-Text row uses the 
SynthID-Text binary tournament backbone and 
applies the budget-matching procedure 
in Appendix~\ref{app:synthid-budget} 
for fair comparison with LUNA.

\subsection{Calibration Details}
\label{app:settings-details}

The context order $k$ is selected from 
$\{2,3,4\}$ separately for LUNA 
and STELA in each language-by-domain setting.
For each algorithm, 
we choose the $k$ that minimizes 
$|\Delta\mathrm{PPL}_{\mathrm{med}}|$ 
on the watermarked outputs.
The selected $k$ values for LUNA and STELA are shown
in
Table~\ref{tab:selected-k};
the two methods agree on the selected order in
$4$ of the $12$ settings.

This selection is applied symmetrically to the
two linguistic methods and is disclosed as part
of the protocol;
the seven remaining baselines do not expose a
POS context-order parameter.
Because the rule optimizes the same statistic
that Table~\ref{tab:main-results} reports,
the resulting headline values are
evaluation-selected, and the bootstrap intervals
of Appendix~\ref{app:bootstrap-cis} summarize
between-setting variation after selection rather
than selection-free uncertainty;
Appendix~\ref{app:k-analysis} reports the
corresponding fixed-order results.
In deployment, where no evaluation outputs are
available, $k$ would instead be chosen once per
language on a held-out calibration split,
for example by maximizing context coverage at a
fixed minimum-count threshold,
before any generation is scored.

\begin{table}[H]
\centering
\small
\setlength{\tabcolsep}{5pt}
\begin{tabular}{@{}l cc cc@{}}
\toprule
 & \multicolumn{2}{c}{\textbf{LUNA}} & \multicolumn{2}{c}{\textbf{STELA}} \\
\cmidrule(lr){2-3} \cmidrule(lr){4-5}
\textbf{Language} & \textbf{Wikipedia} & \textbf{News} & \textbf{Wikipedia} & \textbf{News} \\
\midrule
EN & 3 & 3 & 3 & 4 \\
ZH & 4 & 4 & 2 & 4 \\
KO & 3 & 3 & 4 & 2 \\
JA & 3 & 4 & 2 & 3 \\
DE & 4 & 3 & 2 & 2 \\
AR & 2 & 2 & 2 & 2 \\
\bottomrule
\end{tabular}
\caption{Per-algorithm selected POS context order $k$ 
for the linguistic methods.
Selection minimizes 
$|\Delta\mathrm{PPL}_{\mathrm{med}}|$ within 
each language-by-domain setting. 
LUNA prefers $k\in\{3,4\}$ in 10 of 12 settings, 
while STELA prefers $k=2$ in 7 of 12. 
The per-$k$ comparison appears in 
Appendix~\ref{app:k-analysis}.}
\label{tab:selected-k}
\end{table}

\section{Bootstrap Confidence Intervals for Quality Metrics}
\label{app:bootstrap-cis}

Section~\ref{sec:results-main} reports the bootstrap $95\%$ confidence 
interval for the quality metric 
$|\Delta\mathrm{PPL}_{\mathrm{med}}|$ in Table~\ref{tab:main-results}.
This appendix lists the bootstrap intervals for the 
four remaining quality metrics under the same protocol: 
$1000$ iterations, 
resampling the $12$ language-by-domain settings with 
replacement, seed $42$.
For LUNA and STELA, the intervals 
use the per-algorithm per-setting best context order from 
Table~\ref{tab:selected-k}.

Tables~\ref{tab:bootstrap-cis-lexical} and~\ref{tab:bootstrap-cis-distribution} 
group the four remaining quality metrics by the aspect of distortion they 
capture: lexical structure and next-token-distribution statistics.

\begin{table*}[ht]
\centering
\small
\setlength{\tabcolsep}{4pt}
\begin{minipage}[t]{0.48\textwidth}
\centering
\begin{tabular}{@{}l c c@{}}
\toprule
\textbf{Method} & $|\Delta\textsc{Self-BLEU}|$ & $|\Delta\mathrm{Distinct\text{-}1}|$ \\
\midrule
KGW                  & [0.0045, 0.0081] & [0.0064, 0.0147] \\
EWD                  & [0.0051, 0.0075] & [0.0071, 0.0136] \\
SWEET                & [0.0034, 0.0068] & [0.0041, 0.0096] \\
MorphMark            & [0.0015, 0.0035] & [0.0030, 0.0073] \\
STELA                & [0.0050, 0.0080] & [0.0075, 0.0153] \\
GumbelSoft           & [0.0202, 0.0988] & [0.0222, 0.1159] \\
EXP                  & [0.0251, 0.1229] & [0.0261, 0.1254] \\
SynthID-Text         & [0.0021, 0.0066] & [0.0034, 0.0095] \\
\textbf{LUNA} & \textbf{[0.0010, 0.0022]} & \textbf{[0.0018, 0.0041]} \\
\bottomrule
\end{tabular}
\caption{Bootstrap $95\%$ confidence intervals for the lexical-structure 
quality metrics.
LUNA's upper bound lies strictly below the lower bound of six of eight 
baselines on each metric; 
MorphMark and SynthID-Text are the two methods whose intervals overlap 
LUNA's on both metrics.
Interval endpoints are computed from unrounded bootstrap statistics and 
rounded to four decimals for display.
}
\label{tab:bootstrap-cis-lexical}
\end{minipage}%
\hfill
\begin{minipage}[t]{0.48\textwidth}
\centering
\begin{tabular}{@{}l c c@{}}
\toprule
\textbf{Method} & $|\Delta\mathrm{Surprisal}|$ & $|\Delta\mathrm{Entropy}|$ \\
\midrule
KGW                  & [0.0887, 0.1856] & [0.0435, 0.1152] \\
EWD                  & [0.0796, 0.1604] & [0.0373, 0.0919] \\
SWEET                & [0.0684, 0.1312] & [0.0367, 0.0828] \\
MorphMark            & [0.0228, 0.0667] & [0.0155, 0.0411] \\
STELA                & [0.0874, 0.1658] & [0.0477, 0.0960] \\
GumbelSoft           & [0.0756, 0.2323] & [0.0763, 0.2044] \\
EXP                  & [0.0822, 0.2585] & [0.0903, 0.2455] \\
SynthID-Text         & [0.0320, 0.0730] & [0.0209, 0.0620] \\
\textbf{LUNA} & \textbf{[0.0031, 0.0079]} & \textbf{[0.0067, 0.0174]} \\
\bottomrule
\end{tabular}
\caption{Bootstrap $95\%$ confidence intervals for the 
next-token-distribution quality metrics.
LUNA's upper bound lies strictly below the lower bound of every baseline on 
$|\Delta\mathrm{Surprisal}|$ and of seven of eight baselines on 
$|\Delta\mathrm{Entropy}|$, with MorphMark the only overlap on the latter.
Interval endpoints are computed from unrounded bootstrap statistics and 
rounded to four decimals for display.
}
\label{tab:bootstrap-cis-distribution}
\end{minipage}
\end{table*}

\section{Detection Against Human-Written References}
\label{app:human-negatives}

This appendix reports the full per-setting results
for the human-negative protocol summarized in
Section~\ref{sec:results-human}.

\subsection{Protocol}
\label{app:human-protocol}

The evaluation pipeline scores every record under
two negative classes:
the unwatermarked continuation generated by the
same model from the same prompt~(the main
protocol of Table~\ref{tab:main-results}),
and the gold human continuation supplied by the
dataset record.
Both statistics are computed from the same runs,
so this comparison requires no additional
generation.
The human-negative evaluation uses the same $500$
records per setting, the same score construction,
and, for LUNA and STELA, the same per-setting
context order from Table~\ref{tab:selected-k};
only the negative-class labels differ.
The $5\%$ FPR operating point for TPR is
calibrated on the human negative class of each
setting.
Tables~\ref{tab:vs-human-auroc}
and~\ref{tab:vs-human-tpr} report per-setting
AUROC and TPR at $5\%$ FPR for all nine methods,
and Table~\ref{tab:vs-human-vs-unwm} compares the
twelve-setting means of the two protocols.

\begin{table*}[!ht]
\centering
\small
\setlength{\tabcolsep}{3.2pt}
\begin{tabular}{@{}l cccccc cccccc c@{}}
\toprule
& \multicolumn{6}{c}{\textbf{Wikipedia}} & \multicolumn{6}{c}{\textbf{News}} & \\
\cmidrule(lr){2-7}\cmidrule(lr){8-13}
\textbf{Method} & EN & ZH & KO & JA & DE & AR & EN & ZH & KO & JA & DE & AR & \textbf{Mean} \\
\midrule
KGW           & 0.9992 & 0.9994 & 0.9968 & 0.9993 & 0.9988 & 0.9997 & 0.9979 & 1.0000 & 0.9997 & 1.0000 & 0.9952 & 1.0000 & 0.9988 \\
EWD           & 1.0000 & 1.0000 & 0.9976 & 1.0000 & 0.9981 & 1.0000 & 0.9990 & 1.0000 & 0.9999 & 1.0000 & 0.9936 & 1.0000 & 0.9990 \\
SWEET         & 0.9998 & 0.9981 & 0.9969 & 0.9998 & 0.9995 & 0.9993 & 0.9973 & 1.0000 & 0.9992 & 1.0000 & 0.9937 & 1.0000 & 0.9986 \\
MorphMark     & 0.9921 & 0.9962 & 0.9883 & 0.9868 & 0.9975 & 0.9943 & 0.9894 & 0.9994 & 0.9947 & 0.9983 & 0.9811 & 0.9994 & 0.9931 \\
STELA         & 0.9997 & 1.0000 & 0.9975 & 0.9994 & 0.9970 & 0.9980 & 0.9977 & 1.0000 & 0.9999 & 1.0000 & 0.9967 & 1.0000 & 0.9988 \\
GumbelSoft    & 0.9830 & 0.9821 & 0.9965 & 0.9999 & 0.9900 & 0.9980 & 0.9783 & 0.9901 & 1.0000 & 1.0000 & 0.9613 & 1.0000 & 0.9899 \\
EXP           & 0.9688 & 0.9803 & 0.9943 & 1.0000 & 0.9865 & 1.0000 & 0.9752 & 0.9842 & 1.0000 & 1.0000 & 0.9645 & 1.0000 & 0.9878 \\
SynthID-Text  & 1.0000 & 0.9990 & 0.9942 & 0.9999 & 0.9966 & 0.9978 & 0.9991 & 0.9999 & 0.9996 & 1.0000 & 0.9812 & 1.0000 & 0.9973 \\
\midrule
\textbf{LUNA} & 0.9986 & 0.9970 & 0.9899 & 0.9986 & 0.9955 & 0.9988 & 0.9985 & 0.9999 & 0.9960 & 0.9998 & 0.9804 & 0.9986 & 0.9960 \\
\bottomrule
\end{tabular}
\caption{Detection AUROC with human-written references as the negative
class, for all nine methods across the $12$ language-by-domain
settings~($500$ records per setting).
For LUNA and STELA, values use the per-algorithm, per-setting context order
of Table~\ref{tab:selected-k}.}
\label{tab:vs-human-auroc}
\end{table*}

\begin{table*}[!ht]
\centering
\small
\setlength{\tabcolsep}{3.2pt}
\begin{tabular}{@{}l cccccc cccccc c@{}}
\toprule
& \multicolumn{6}{c}{\textbf{Wikipedia}} & \multicolumn{6}{c}{\textbf{News}} & \\
\cmidrule(lr){2-7}\cmidrule(lr){8-13}
\textbf{Method} & EN & ZH & KO & JA & DE & AR & EN & ZH & KO & JA & DE & AR & \textbf{Mean} \\
\midrule
KGW           & 0.998 & 0.998 & 0.992 & 0.998 & 0.998 & 0.998 & 0.992 & 1.000 & 0.998 & 1.000 & 0.979 & 1.000 & 0.996 \\
EWD           & 1.000 & 1.000 & 0.996 & 1.000 & 0.994 & 1.000 & 0.996 & 1.000 & 1.000 & 1.000 & 0.980 & 1.000 & 0.997 \\
SWEET         & 0.998 & 0.998 & 0.996 & 0.998 & 0.998 & 0.996 & 0.982 & 1.000 & 0.998 & 1.000 & 0.978 & 1.000 & 0.995 \\
MorphMark     & 0.974 & 0.988 & 0.964 & 0.937 & 0.986 & 0.970 & 0.970 & 0.998 & 0.980 & 0.996 & 0.960 & 0.998 & 0.977 \\
STELA         & 1.000 & 1.000 & 0.990 & 1.000 & 0.992 & 0.994 & 0.992 & 1.000 & 1.000 & 1.000 & 0.986 & 1.000 & 0.996 \\
GumbelSoft    & 0.960 & 0.958 & 0.996 & 1.000 & 0.980 & 0.998 & 0.958 & 0.974 & 1.000 & 1.000 & 0.926 & 1.000 & 0.979 \\
EXP           & 0.948 & 0.958 & 0.994 & 1.000 & 0.978 & 1.000 & 0.962 & 0.972 & 1.000 & 1.000 & 0.940 & 1.000 & 0.979 \\
SynthID-Text  & 1.000 & 0.996 & 0.986 & 1.000 & 0.984 & 0.994 & 0.998 & 1.000 & 1.000 & 1.000 & 0.950 & 1.000 & 0.992 \\
\midrule
\textbf{LUNA} & 0.996 & 0.992 & 0.958 & 0.992 & 0.988 & 0.992 & 0.990 & 1.000 & 0.980 & 1.000 & 0.958 & 0.998 & 0.987 \\
\bottomrule
\end{tabular}
\caption{Detection TPR at $5\%$ FPR with human-written references as the
negative class;
the operating point is calibrated on the human negatives of each setting.
Same protocol as Table~\ref{tab:vs-human-auroc};
values are rounded to three decimals.}
\label{tab:vs-human-tpr}
\end{table*}

\begin{table*}[!ht]
\centering
\setlength{\tabcolsep}{4pt}
\begin{tabular}{@{}l cc cc@{}}
\toprule
& \multicolumn{2}{c}{\textbf{AUROC}} & \multicolumn{2}{c}{\textbf{TPR@5\%}} \\
\cmidrule(lr){2-3}\cmidrule(lr){4-5}
\textbf{Method} & Unwatermarked & Human & Unwatermarked & Human \\
\midrule
KGW          & 0.9982 & 0.9988 & 0.9952 & 0.9959 \\
EWD          & 0.9990 & 0.9990 & 0.9972 & 0.9972 \\
SWEET        & 0.9985 & 0.9986 & 0.9950 & 0.9951 \\
MorphMark    & 0.9902 & 0.9931 & 0.9643 & 0.9767 \\
STELA        & 0.9982 & 0.9988 & 0.9953 & 0.9962 \\
GumbelSoft   & 0.9899 & 0.9899 & 0.9778 & 0.9792 \\
EXP          & 0.9876 & 0.9878 & 0.9777 & 0.9793 \\
SynthID-Text & 0.9972 & 0.9973 & 0.9928 & 0.9923 \\
\textbf{LUNA} & 0.9959 & 0.9960 & 0.9868 & 0.9870 \\
\bottomrule
\end{tabular}
\caption{Twelve-setting mean detection under the two negative-class
protocols:
unwatermarked outputs of the same model~(Unwatermarked) and
human-written references~(Human).
For every method the AUROC difference is at most $0.003$,
and the TPR difference is at most $0.002$ for all methods except
MorphMark~($+0.012$).}
\label{tab:vs-human-vs-unwm}
\end{table*}

\subsection{Behavior Across Methods}
\label{app:human-behavior}

Detection remains near-saturated under the
human-negative protocol.
EWD, KGW, STELA, SWEET, and SynthID-Text reach
twelve-setting mean AUROC between $0.9973$ and
$0.9990$, and LUNA reaches $0.9960$.
In the Chinese, Japanese, and Arabic news
settings, at least five methods attain AUROC
$\ge 0.9999$,
so per-setting rankings within this band reflect
noise in the last decimals rather than a stable
separation, which is the same saturation reading
that Section~\ref{sec:results-main} applies to
the main protocol.
MorphMark, GumbelSoft, and EXP occupy the lower
end of the band, with means between $0.9878$ and
$0.9931$, and account for the widest per-setting
spreads.

LUNA's per-setting AUROC lies within $0.005$ of
the best method of that setting in ten of the
twelve settings and within $0.017$ in all twelve;
its lowest per-setting value is $0.9804$ on
German news.
German news is the hardest setting for every
method: each of the nine methods records its
per-setting minimum there, with values between
$0.9613$ and $0.9967$.
The English and Chinese Wikipedia settings are
hard mainly for the Gumbel-family methods,
where GumbelSoft and EXP fall to
$0.9688$--$0.9830$ while the remaining methods
stay above $0.988$.

\subsection{Agreement Between the Two Protocols}
\label{app:human-agreement}

Table~\ref{tab:vs-human-vs-unwm} compares the two
protocols at the twelve-setting mean.
For every method the absolute AUROC difference is
at most $0.003$, and the TPR difference is at
most $0.002$ for all methods except
MorphMark~($+0.012$),
the method with the lowest TPR at $5\%$ FPR under
either protocol.
This agreement indicates that detector null
behavior calibrated on unwatermarked outputs
transfers to the evaluated human references,
so the headline comparison of
Table~\ref{tab:main-results} is not an artifact
of the negative-class choice.

\subsection{Why the Main Protocol Uses Unwatermarked Negatives}
\label{app:human-rationale}

Three considerations favor unwatermarked
negatives as the headline protocol.
First, it is the standard reporting convention of
the primary baselines, which keeps the reported
numbers directly comparable to prior
work~\citep{kirchenbauer2023watermark,kuditipudi2024robust,lu-etal-2024-entropy,lee-etal-2024-wrote,dathathri2024scalable}.
Second, it matches the provenance decision that
watermarking operationalizes, namely whether a
text came from the provider's watermarked
deployment rather than from the same model
without the watermark,
and it isolates watermark-specific evidence by
matching the two classes on generator, prompts,
decoding configuration, and length statistics.
Third, human references vary in ways unrelated to
the watermark, such as unusually short or
template-like continuations,
which at four-decimal resolution can perturb a
per-setting AUROC through a handful of outlier
records.
The human-negative results complement rather than
replace the main protocol,
and, as Section~\ref{sec:results-human} notes,
they do not imply a universal fixed-threshold
false-positive rate for arbitrary human corpora.

\section{Design and Analysis of SynthID-Text-Entropy}
\label{app:synthid-entropy}

This appendix defines the 
SynthID-Text-Entropy used in
Section~\ref{sec:results-ablations}.
This variant is not a previously published watermark.
It is a diagnostic baseline 
that asks whether a model-derived
entropy signal can replace 
the external linguistic signal used by LUNA.

\subsection{Design Rationale}
\label{app:sidte-design}

LUNA combines three ingredients: 
a non-distortionary SynthID-Text tournament
backbone, 
a prefix-measurable adaptive schedule derived 
from POS-context
entropy, 
and a detector that reconstructs the same linguistic schedule without
language-model forward passes.
STELA tests the value of replacing a 
distortionary linguistic watermark with 
a non-distortionary tournament backbone.
SynthID-Text tests the value of adding a 
linguistic schedule to a tournament sampler.
SynthID-Text-Entropy tests a third question: 
whether model-side entropy can
play the role that POS-context entropy plays in LUNA.

SynthID-Text-Entropy
keeps the SynthID-Text tournament family
and the budget-matching
procedure used for the SynthID-Text baseline,
and it consumes token-level model entropy at both
stages, exactly where LUNA consumes \(\lambda(c_t)\).
At generation time, the predictive entropy of the
next-token distribution \(p_t\) sets the
tournament depth \(m_t\) through a tier schedule
whose thresholds are calibrated so that the
expected candidate budget matches LUNA's;
at detection time, the verifier recomputes
\(p_t\) with language-model forward passes to
recover the same per-token entropies and
therefore the same schedule.
The adaptive signal is thus model-side at
insertion and at verification alike, which makes
this variant the direct mirror of LUNA's
corpus-side design.
This choice creates a strong model-aware comparison point.
It also removes model-free detection,
since the verifier must run a language
model to obtain per-token entropy values.

\subsection{Budget Matching with LUNA}
\label{app:sidte-budget}

We match the expected tournament budget of
SynthID-Text-Entropy to LUNA using
the same procedure as Appendix~\ref{app:synthid-budget}.
Let \(B=\mathbb{E}[2^{m_t}]\) 
denote the expected tournament budget induced by
the LUNA depth ladder under the calibration proportions
\((p_{\mathrm{low}},p_{\mathrm{mid}},p_{\mathrm{high}})\).
The SynthID-Text-Entropy configuration uses 
the corresponding budget-matched
SynthID-Text tournament schedule, 
so the comparison is not driven by a larger
average tournament budget.

\subsection{Comparison and Practical Implications}
\label{app:sidte-practical}

At the \(12\)-setting mean, SynthID-Text-Entropy attains AUROC \(0.9960\) and
\(|\Delta\mathrm{PPL}_{\mathrm{med}}|=0.0787\), 
while LUNA attains AUROC
\(0.9959\) and \(|\Delta\mathrm{PPL}_{\mathrm{med}}|=0.0447\).
Detection is effectively indistinguishable
at this aggregate level: the AUROC
gap is \(0.0001\) in favor of SynthID-Text-Entropy, 
and the TPR@5\% gap is
\(0.0007\).
On quality preservation, 
LUNA improves four of the five reported quality
metrics by factors of \(1.59\times\) to \(1.76\times\), while
SynthID-Text-Entropy is \(1.09\times\) better on
\(|\Delta\mathrm{Distinct\text{-}1}|\).

The comparison clarifies the deployment trade-off.
Model entropy supplies a powerful adaptive signal, 
yet it requires
language-model forward passes at verification time.
This dependence creates serving cost, 
version coupling, and weaker
third-party verifiability when the generator 
or an appropriate surrogate is
not available.
LUNA reaches the same detection regime without 
this dependence and preserves
quality better on most reported metrics.

\section{Detection-Quality Trade-off}
\label{app:trade-off}

This appendix visualizes the per-setting structure 
that underlies the 
aggregate detection and quality results in Section~\ref{sec:results-main}.
We characterize the trade-off space through 
three complementary views: 
the Pareto frontier~(\ref{app:pareto}), 
the per-setting sweet-spot distribution~(\ref{app:sweet-spot}), and 
the per-baseline multi-metric quality advantage~(\ref{app:multi-metric}).

\subsection{Pareto Frontier of the Detection--Quality Trade-off}
\label{app:pareto}

\begin{figure*}[ht!]
\centering
\includegraphics[width=\textwidth]{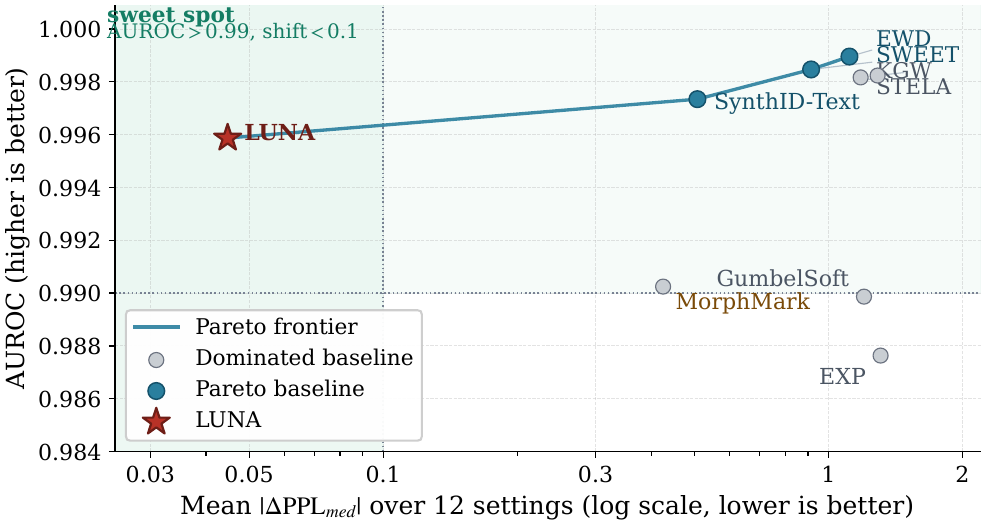}
\caption{Pareto frontier of the detection-quality 
trade-off, with AUROC on 
the vertical axis and $|\Delta\mathrm{PPL}_{\mathrm{med}}|$ on the 
horizontal axis~(log scale), 
averaged over the $12$ language-by-domain 
settings.
Four of nine methods are 
Pareto-optimal~(filled markers, connected by the 
frontier); 
the other five are dominated~(gray markers).
The shaded sweet-spot region in the 
upper-left corner marks 
AUROC $>0.99$ and shift $<0.1$; LUNA is the only method that enters it.}
\label{fig:pareto-front}
\end{figure*}

Figure~\ref{fig:pareto-front} plots the 
$12$-setting mean of AUROC against 
$|\Delta\mathrm{PPL}_{\mathrm{med}}|$ on a 
logarithmic horizontal axis.
Four of the nine methods are Pareto-optimal: 
EWD, SWEET, SynthID-Text, and 
LUNA; 
the remaining five~(KGW, STELA, MorphMark, EXP, GumbelSoft) are 
dominated by some method that achieves 
both better detection and 
lower distortion.

LUNA occupies the left endpoint of the Pareto front.
The nearest Pareto neighbor, SynthID-Text, 
sits at $|\Delta\mathrm{PPL}_{\mathrm{med}}|=0.463$ 
with AUROC $0.9972$; 
moving to LUNA reduces $|\Delta\mathrm{PPL}_{\mathrm{med}}|$ by a factor of 
$10.4\times$ at an AUROC cost of $0.0013$.
The shaded sweet-spot region marks 
the operating regime where 
AUROC $>0.99$ and $|\Delta\mathrm{PPL}_{\mathrm{med}}|<0.1$; 
LUNA is the only Pareto-optimal method inside this region, and the only 
method of the nine to enter it at the $12$-setting mean.

\subsection{Per-Setting Sweet-Spot Distribution}
\label{app:sweet-spot}

\begin{figure*}[ht!]
\centering
\includegraphics[width=\textwidth]{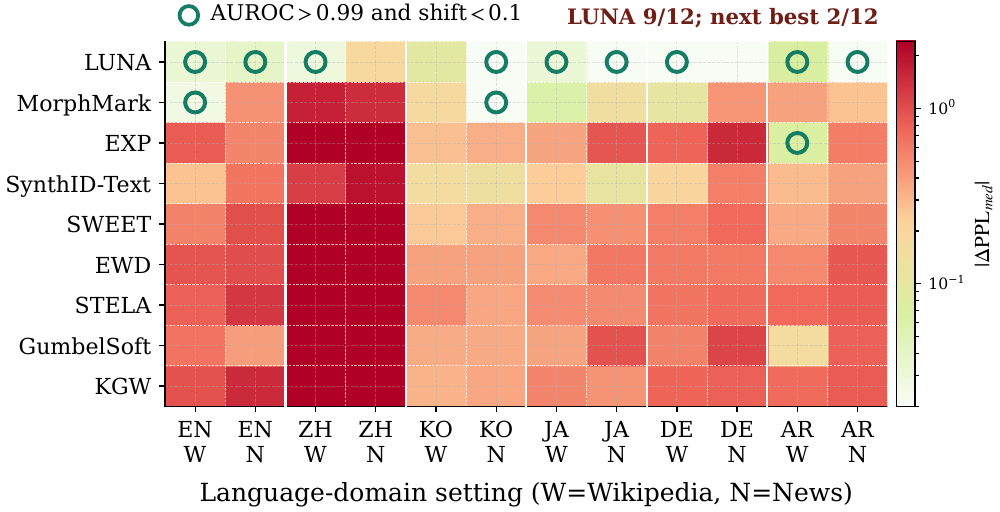}
\caption{Per-setting $|\Delta\mathrm{PPL}_{\mathrm{med}}|$ on a logarithmic 
scale for all nine methods across 
the $12$ language-by-domain settings.
Rows are ordered by the number of 
sweet-spot cells~(green circles, marking 
AUROC $>0.99$ and shift $<0.1$).
LUNA enters the sweet-spot in $9$ of $12$ settings; 
the next-best baseline~(MorphMark) enters it in $2$ of $12$ settings.}
\label{fig:sweet-spot-heatmap}
\end{figure*}

The aggregate sweet-spot finding holds 
at the per-setting level.
Figure~\ref{fig:sweet-spot-heatmap} colors 
each~(method, language-domain) 
cell by $|\Delta\mathrm{PPL}_{\mathrm{med}}|$ 
on a logarithmic scale; 
green circles mark the cells that jointly satisfy 
AUROC $>0.99$ and 
$|\Delta\mathrm{PPL}_{\mathrm{med}}|<0.1$.
LUNA reaches the sweet-spot in $9$ of $12$ settings.

\subsection{Per-Baseline Multi-Metric Quality Advantage}
\label{app:multi-metric}

\begin{figure*}[ht!]
\centering
\includegraphics[width=\textwidth]{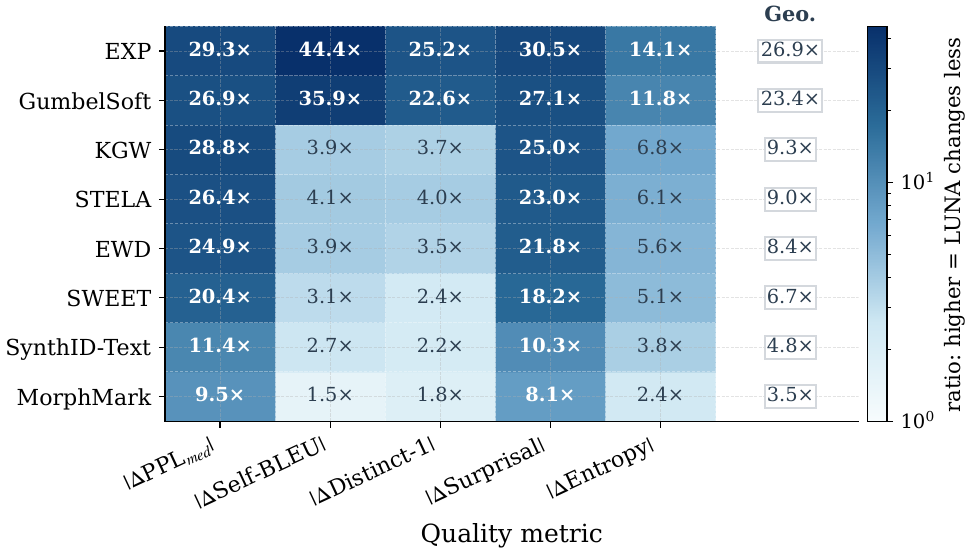}
\caption{Per-baseline multi-metric quality advantage of LUNA, computed as 
the ratio between each baseline's $12$-setting mean and LUNA's on the five 
quality metrics.
Cells with ratio $>1$ indicate LUNA changes the metric by a smaller amount.
The rightmost column reports the geometric mean across the five metrics.
LUNA holds a uniform advantage on every~(baseline, metric) cell of the 
table across the eight main baselines.
The SynthID-Text-Entropy ablation in Section~\ref{sec:results-ablations} is 
not shown here and is the one comparison in which LUNA does not dominate on 
every metric.}
\label{fig:quality-ratio}
\end{figure*}

Figure~\ref{fig:quality-ratio} extends the comparison 
from the single 
$|\Delta\mathrm{PPL}_{\mathrm{med}}|$ axis 
to all five quality metrics 
simultaneously.
Each cell reports the ratio of the baseline's mean distortion to LUNA's; 
the rightmost column reports the geometric mean 
across the five metrics.
LUNA's geometric-mean advantage over the closest
baseline~(MorphMark) is 
$3.5\times$, and the advantage exceeds 
an order of magnitude against KGW, 
STELA, EWD, GumbelSoft, and EXP.
The largest single-metric ratios are observed for
$|\Delta\mathrm{PPL}_{\mathrm{med}}|$ and $|\Delta\mathrm{Surprisal}|$,
both of which are governed by the realized 
next-token probability under the reference model.
The lexical-structure metrics~($|\Delta\textsc{Self-BLEU}|$, $|\Delta\mathrm{Distinct\text{-}1}|$) 
also show consistent positive gains 
at smaller magnitudes, suggesting that the 
quality preservation remains robust across diverse 
forms of distortion rather than being 
attributable to improvements in a single metric alone.

\section{Behavior Across Experimental Axes}
\label{app:axes}

This appendix examines whether the aggregate behavior 
in Section~\ref{sec:results} is uniform across 
the experimental axes.
We report per-language ranks~(\ref{app:per-language}) and
per-domain ranks~(\ref{app:per-domain}).

\subsection{Per-Language Behavior}
\label{app:per-language}

Table~\ref{tab:per-language-ranks} reports 
LUNA's rank among the nine methods 
on each of the seven metrics, separately for each language.
Ranks are computed on the per-language mean 
over Wikipedia and news.
LUNA holds rank $1$ on $|\Delta\mathrm{PPL}_{\mathrm{med}}|$ and 
$|\Delta\mathrm{Surprisal}|$ in all six languages, 
on $|\Delta\textsc{Self-BLEU}|$ and $|\Delta\mathrm{Entropy}|$ in five of six, 
and on $|\Delta\mathrm{Distinct\text{-}1}|$ in three of six.
The quality advantage is consistently preserved 
across typologically diverse languages, 
including analytic English, isolating Chinese, 
agglutinative Korean and Japanese, fusional German, 
and Semitic-templatic Arabic.
Detection ranks range from $3$ to $8$,
while remaining entirely within the saturated AUROC
regime identified in Section~\ref{sec:results-main}.
In particular, Arabic, the language with the
narrowest threshold spread in
Table~\ref{tab:lambda-spread},
ranks first on all five quality metrics,
and both Arabic settings satisfy the joint
criterion AUROC $>0.99$ and
$|\Delta\mathrm{PPL}_{\mathrm{med}}|<0.1$ in
Figure~\ref{fig:sweet-spot-heatmap},
so the quality advantage is not confined to
languages with wide $\lambda$ spreads.

\begin{table*}[!ht]
\centering
\small
\setlength{\tabcolsep}{3.0pt}
\begin{tabular}{@{}l c c c c c c c@{}}
\toprule
\textbf{Lang.} & \textbf{AUROC} & \textbf{TPR@5\%}
& \(\boldsymbol{|\Delta\mathrm{PPL}_{\mathrm{med}}|}\)
& \(\boldsymbol{|\Delta\textsc{Self-BLEU}|}\)
& \(\boldsymbol{|\Delta\mathrm{Distinct\text{-}1}|}\)
& \(\boldsymbol{|\Delta\mathrm{Surprisal}|}\)
& \(\boldsymbol{|\Delta\mathrm{Entropy}|}\) \\
\midrule
EN & 3 & 5 & \textbf{1} & \textbf{1} & 2          & \textbf{1} & \textbf{1} \\
ZH & 6 & 6 & \textbf{1} & 2          & \textbf{1} & \textbf{1} & \textbf{1} \\
KO & 8 & 8 & \textbf{1} & \textbf{1} & 5          & \textbf{1} & \textbf{1} \\
JA & 8 & 8 & \textbf{1} & \textbf{1} & \textbf{1} & \textbf{1} & 2          \\
DE & 6 & 6 & \textbf{1} & \textbf{1} & 3          & \textbf{1} & \textbf{1} \\
AR & 7 & 8 & \textbf{1} & \textbf{1} & \textbf{1} & \textbf{1} & \textbf{1} \\
\bottomrule
\end{tabular}
\caption{LUNA's rank out of \(9\) methods on each metric, per language,
computed on the per-language mean over Wikipedia and news.
Bold entries indicate rank \(1\).}
\label{tab:per-language-ranks}
\end{table*}

\subsection{Per-Domain Behavior}
\label{app:per-domain}

Appendix~\ref{app:per-language} reports per-language ranks.
This subsection reports the same rank summary 
for the two text domains~(Table~\ref{tab:per-domain-ranks}).
LUNA is rank $1$ on $|\Delta\mathrm{PPL}_{\mathrm{med}}|$, 
$|\Delta\mathrm{Surprisal}|$, and $|\Delta\mathrm{Entropy}|$ in both 
Wikipedia and news; 
on the two lexical-structure metrics it alternates between rank $1$ and rank 
$2$ across domains.
Detection rank is $6$ of $9$ in both domains, inside the saturated AUROC 
band.
The trade-off profile is symmetric across the two domains.

\begin{table*}[!ht]
\centering
\small
\setlength{\tabcolsep}{3.5pt}
\begin{tabular}{@{}l c c c c c c c@{}}
\toprule
\textbf{Domain} & \textbf{AUROC} & \textbf{TPR@5\%}
& \(\boldsymbol{|\Delta\mathrm{PPL}_{\mathrm{med}}|}\)
& \(\boldsymbol{|\Delta\textsc{Self-BLEU}|}\)
& \(\boldsymbol{|\Delta\mathrm{Distinct\text{-}1}|}\)
& \(\boldsymbol{|\Delta\mathrm{Surprisal}|}\)
& \(\boldsymbol{|\Delta\mathrm{Entropy}|}\) \\
\midrule
Wikipedia & 6 & 6 & \textbf{1} & \textbf{1} & 2          & \textbf{1} & \textbf{1} \\
News      & 6 & 6 & \textbf{1} & 2          & \textbf{1} & \textbf{1} & \textbf{1} \\
\bottomrule
\end{tabular}
\caption{LUNA's rank out of \(9\) methods on each metric, 
per domain,
computed on the \(6\)-language mean within 
each domain.
Bold entries indicate rank \(1\).}
\label{tab:per-domain-ranks}
\end{table*}

\section{Context-Order Analysis}
\label{app:k-analysis}

This appendix examines the POS context-order hyperparameter \(k\) for the two
linguistic methods, LUNA and STELA.
Both methods consume the same corpus-estimated POS-context signal, yet they
use it in different sampling mechanisms.
LUNA turns \(\lambda(c_t)\) into tournament depth, whereas STELA turns the same
signal into green-list bias and detector weights.
We therefore restrict this analysis to LUNA and STELA, since the goal is to
understand how linguistic context length interacts with the two
linguistic-signal mechanisms.

\subsection{\texorpdfstring{\(k\)}{k}-Stratified Comparison}
\label{app:k-stratified}

Tables~\ref{tab:ablation-k2}--\ref{tab:ablation-k4} report the fixed-\(k\)
means for LUNA and STELA at \(k\in\{2,3,4\}\).
These fixed-order results underlie the fixed-order robustness summary in
Section~\ref{sec:results-main} and involve no per-setting selection.
At every fixed order, LUNA records smaller shifts than STELA on all five
quality metrics while maintaining AUROC between \(0.9947\) and \(0.9958\);
STELA's perplexity shifts are \(9.31\times\) to \(13.59\times\) as large
across the three orders.
The comparison shows that LUNA preserves the quality advantage across context
orders, while the optimal order varies by method and setting.
This pattern motivates the setting-level selection rule in
Table~\ref{tab:selected-k}.

\begin{table*}[!ht]
\centering
\small
\setlength{\tabcolsep}{4pt}
\begin{tabular}{@{}l c c c c c c c@{}}
\toprule
\textbf{Method} & \textbf{AUROC} & \textbf{TPR@5\%}
& \( |\Delta\mathrm{PPL}_{\mathrm{med}}| \)
& \( |\Delta\textsc{Self-BLEU}| \)
& \( |\Delta\mathrm{Distinct\text{-}1}| \)
& \( |\Delta\mathrm{Surprisal}| \)
& \( |\Delta\mathrm{Entropy}| \) \\
\midrule
LUNA  & 0.9950 & 0.9818 & 0.1227 & 0.0013 & 0.0032 & 0.0154 & 0.0141 \\
STELA & 0.9985 & 0.9955 & 1.2225 & 0.0064 & 0.0114 & 0.1297 & 0.0755 \\
\bottomrule
\end{tabular}
\caption{Fixed-order comparison between LUNA and STELA at \(k=2\), averaged
over the \(12\) language-by-domain settings.}
\label{tab:ablation-k2}
\end{table*}

\begin{table*}[!ht]
\centering
\small
\setlength{\tabcolsep}{4pt}
\begin{tabular}{@{}l c c c c c c c@{}}
\toprule
\textbf{Method} & \textbf{AUROC} & \textbf{TPR@5\%}
& \( |\Delta\mathrm{PPL}_{\mathrm{med}}| \)
& \( |\Delta\textsc{Self-BLEU}| \)
& \( |\Delta\mathrm{Distinct\text{-}1}| \)
& \( |\Delta\mathrm{Surprisal}| \)
& \( |\Delta\mathrm{Entropy}| \) \\
\midrule
LUNA  & 0.9958 & 0.9862 & 0.0999 & 0.0012 & 0.0029 & 0.0111 & 0.0142 \\
STELA & 0.9991 & 0.9978 & 1.3573 & 0.0066 & 0.0109 & 0.1451 & 0.0862 \\
\bottomrule
\end{tabular}
\caption{Fixed-order comparison between LUNA and STELA at \(k=3\), averaged
over the \(12\) language-by-domain settings.}
\label{tab:ablation-k3}
\end{table*}

\begin{table*}[!ht]
\centering
\small
\setlength{\tabcolsep}{4pt}
\begin{tabular}{@{}l c c c c c c c@{}}
\toprule
\textbf{Method} & \textbf{AUROC} & \textbf{TPR@5\%}
& \( |\Delta\mathrm{PPL}_{\mathrm{med}}| \)
& \( |\Delta\textsc{Self-BLEU}| \)
& \( |\Delta\mathrm{Distinct\text{-}1}| \)
& \( |\Delta\mathrm{Surprisal}| \)
& \( |\Delta\mathrm{Entropy}| \) \\
\midrule
LUNA  & 0.9947 & 0.9821 & 0.1376 & 0.0041 & 0.0048 & 0.0189 & 0.0175 \\
STELA & 0.9983 & 0.9952 & 1.2805 & 0.0045 & 0.0108 & 0.1386 & 0.0779 \\
\bottomrule
\end{tabular}
\caption{Fixed-order comparison between LUNA and STELA at \(k=4\), averaged
over the \(12\) language-by-domain settings.}
\label{tab:ablation-k4}
\end{table*}

\subsection{Context-Order Selection Patterns}
\label{app:k-pattern}

The setting-level selections in Table~\ref{tab:selected-k} show that LUNA and
STELA prefer different context lengths.
LUNA selects \(k\in\{3,4\}\) in \(10\) of \(12\) settings, whereas STELA
selects \(k=2\) in \(7\) of \(12\) settings.
The two methods agree on the selected \(k\) in only \(4\) of \(12\) settings.
This difference suggests that the same linguistic signal interacts differently
with the sampling mechanism.
LUNA can exploit longer POS contexts through depth modulation, while STELA
often prefers shorter contexts when the signal drives a distortionary
green-list bias.
Two mechanism-level differences are consistent with this pattern.
First, LUNA quantizes \(\lambda(c_t)\) into three depth tiers, so a change in
\(\lambda\) that stays within a tier leaves the insertion depth unchanged,
which makes depth scheduling comparatively tolerant of the sparser,
higher-variance estimates that longer contexts produce.
Second, longer contexts trigger more order backoff under the minimum-count
threshold of Algorithm~\ref{alg:lookup}, and backoff interacts differently
with a continuous green-list bias than with discrete depth tiers.

\section{Linguistic Behavior of \texorpdfstring{$\lambda$}{lambda} Across Languages}
\label{app:lambda-typology}

This appendix expands the linguistic intuition behind 
the normalized next-tag entropy $\lambda(c)$.
We describe the kind of POS context that LUNA tends 
to mark as low or high $\lambda$ in each language, 
and we report the spread of $\tau_2-\tau_1$ measured 
on the calibration corpus at the 
selected primary order from Table~\ref{tab:selected-k}.
The spread is the gap between the frequency-weighted 
25th and 75th percentiles of $\lambda$ in that language; 
it summarizes how widely $\lambda$ varies 
across positions, 
and therefore how often LUNA chooses the deepest 
tier rather than the shallowest.

\paragraph{Why the spread of $\lambda$ matters.}
LUNA applies the deep tournament tier 
only at positions whose $\lambda$ value exceeds $\tau_2$.
A wider spread therefore means that the deep tier 
is reserved for positions that are genuinely more 
uncertain than typical positions in the same language, 
rather than being applied uniformly.
A narrow spread means that most positions sit close
to a common $\lambda$ value.
Numerical spread and realized tier occupancy are
distinct, however:
because the thresholds are language-specific
frequency-weighted percentiles, the calibration
distribution populates the three tiers at the
proportions
$(p_{\mathrm{low}},p_{\mathrm{mid}},p_{\mathrm{high}})$
used by the budget match of
Appendix~\ref{app:synthid-budget},
and close cutpoints do not by themselves imply
that a single depth dominates.
A narrow spread instead means that the tier
boundaries separate positions whose estimated
uncertainties are numerically similar,
so tier membership is more sensitive to
estimation noise in $\lambda$.
The spread is a property of the language and its tagger,
not of the watermark;
the watermark only consumes this signal.

\paragraph{Per-language behavior.}

\textit{Korean.}
Korean is agglutinative with overt particles 
and verbal endings, 
and the Sejong tagset distinguishes 
nominal markers, case markers, and verb-ending morphemes.
A POS context that ends with a topic marker 
can be followed by many tag types depending on whether 
the sentence continues with a verb phrase, 
a coordinated clause, or an embedded clause.
A POS context that ends with a clausal final ending 
is far more constrained.
The two regimes are reflected in a wide $\lambda$ spread.

\textit{German.}
German shows fusional case-and-number agreement 
and verb-second main-clause syntax~\citep{haider2010syntax,vikner1995verb}.
The position immediately after a fronted constituent 
in a main clause is fixed to a finite verb.
The position after a finite verb is much more open, 
since it can host a subject, an adverb, 
or a nominal complement depending on the construction.
This contrast between syntactically constrained 
verb-second positions and freer post-verb positions 
yields a wide $\lambda$ spread, close to Korean.

\textit{English.}
English has light inflection and rigid 
SVO word order~\citep{quirk1985comprehensive}.
Determiner-adjective contexts almost always continue 
with a noun, 
while preposition-noun contexts can be followed 
by several functional categories.
The result is a moderate spread.

\textit{Japanese.}
Japanese is agglutinative like Korean, 
yet writing mixes hiragana, katakana, and kanji, 
and SudachiPy splits compound nouns 
into morphemes~\citep{takaoka-etal-2018-sudachi}.
This segmentation flattens distinctions among many 
nominal contexts, 
so $\lambda$ varies less across positions than in 
Korean despite a comparable underlying morphology.

\textit{Chinese.}
Mandarin Chinese is isolating and uses 
few overt grammatical markers~\citep{li-thompson-1981}.
Most POS contexts allow a similar set of continuations, 
dominated by nouns and verbs, so $\lambda$ 
remains close to its language-level mean.

\textit{Arabic.}
Arabic combines templatic root-and-pattern morphology 
with rich agreement and an abjad script~\citep{mccarthy1981prosodic,ryding2005reference}.
The CAMeL Tools tagger emits fine-grained tags
that already encode much of this morphological information,
so consecutive tags carry a high mutual constraint.
Combined with the small selected order $k=2$,
the numerical spread between the percentile
thresholds is comparatively small,
although the percentile construction still
assigns positions to all three tiers.

\paragraph{Measured spread.}
Table~\ref{tab:lambda-spread} 
reports $\tau_1$, $\tau_2$, and the 
spread $\tau_2-\tau_1$ averaged over the Wikipedia 
and news calibration corpora at the selected primary order.
The order from widest to narrowest spread is Korean, 
German, English, Japanese, Chinese, Arabic.
This ordering matches the per-language narrative 
above and supports the interpretation of $\lambda$ 
as a linguistic capacity signal.

\begin{table}[H]
\centering
\setlength{\tabcolsep}{6pt}
\begin{tabular}{@{}l c c c@{}}
\toprule
\textbf{Lang.} & $\tau_1$ & $\tau_2$ & $\tau_2-\tau_1$ \\
\midrule
KO & 0.374 & 0.618 & 0.244 \\
DE & 0.520 & 0.748 & 0.229 \\
EN & 0.558 & 0.696 & 0.137 \\
JA & 0.522 & 0.640 & 0.118 \\
ZH & 0.553 & 0.663 & 0.111 \\
AR & 0.454 & 0.519 & 0.065 \\
\bottomrule
\end{tabular}
\caption{Frequency-weighted 25th and 75th percentile 
thresholds of $\lambda$ for LUNA at the 
selected primary order from Table~\ref{tab:selected-k},
averaged over Wikipedia and news.
The spread $\tau_2-\tau_1$ summarizes how widely 
the linguistic capacity signal varies across positions 
in each language.
Thresholds and spreads are computed from unrounded 
values and rounded independently for display.
}
\label{tab:lambda-spread}
\end{table}

\end{document}